\documentclass{article}

\usepackage[preprint]{neurips_2025}

\usepackage[utf8]{inputenc} % allow utf-8 input
\usepackage[T1]{fontenc}    % use 8-bit T1 fonts
\usepackage{amsmath}
\usepackage{amsthm}

\usepackage{hyperref}       % hyperlinks
\usepackage{url}            % simple URL typesetting
\usepackage{booktabs}       % professional-quality tables
\usepackage{amsfonts}       % blackboard math symbols
\usepackage{nicefrac}       % compact symbols for 1/2, etc.
\usepackage{microtype}      % microtypography
\usepackage{xcolor}         % colors
\usepackage[table]{xcolor}
\usepackage{graphicx}
\usepackage{diagbox}
\usepackage{url}
\usepackage{booktabs}
\usepackage{amssymb}
\usepackage{bbding}
\usepackage{pifont}
\usepackage{wasysym}
\usepackage{utfsym}
\usepackage{fontawesome}
\usepackage{appendix}
\usepackage{multirow}
\usepackage{wrapfig}
\usepackage{enumitem}
\usepackage[breakable]{tcolorbox}
\usepackage{tabularx}

\title{A Creative Agent is Worth a 64-Token Template}

\author{Ruixiao Shi\textsuperscript{\rm 1,2}\thanks{Equal contribution}\quad Fu Feng\textsuperscript{\rm 1,2}\footnotemark[1]\quad Yucheng Xie\textsuperscript{\rm 1,2}\quad Xu Yang\textsuperscript{\rm 1,2}\quad Jing Wang\textsuperscript{\rm 1,2}\thanks{Corresponding authors}\quad Xin Geng\textsuperscript{\rm 1,2}\footnotemark[2]\\
\textsuperscript{\rm 1}School of Computer Science and Engineering, Southeast University, Nanjing, China\\
\textsuperscript{\rm 2}Key Laboratory of New Generation Artificial Intelligence Technology and Its Interdisciplinary \\Applications (Southeast University), Ministry of Education, China\\
{\texttt \{eric\_xiao, fufeng, xieyc, xuyang\_palm, wangjing91, xgeng\}@seu.edu.cn}\\
}

\begin{document}

\maketitle
% \vspace{-5mm}
% \begin{center}
%     \includegraphics[width=\textwidth]{imgs/CreTemp_teasor.pdf}
% \end{center}
% \vspace{-6mm}
% \begin{center}
% {\small
% \textbf{Figure 1:} Distribution-Conditional Generation...
% }
% \end{center}

\begin{figure}[ht]
  \centering
  \vspace{-0.3in}
  \includegraphics[width=\linewidth]{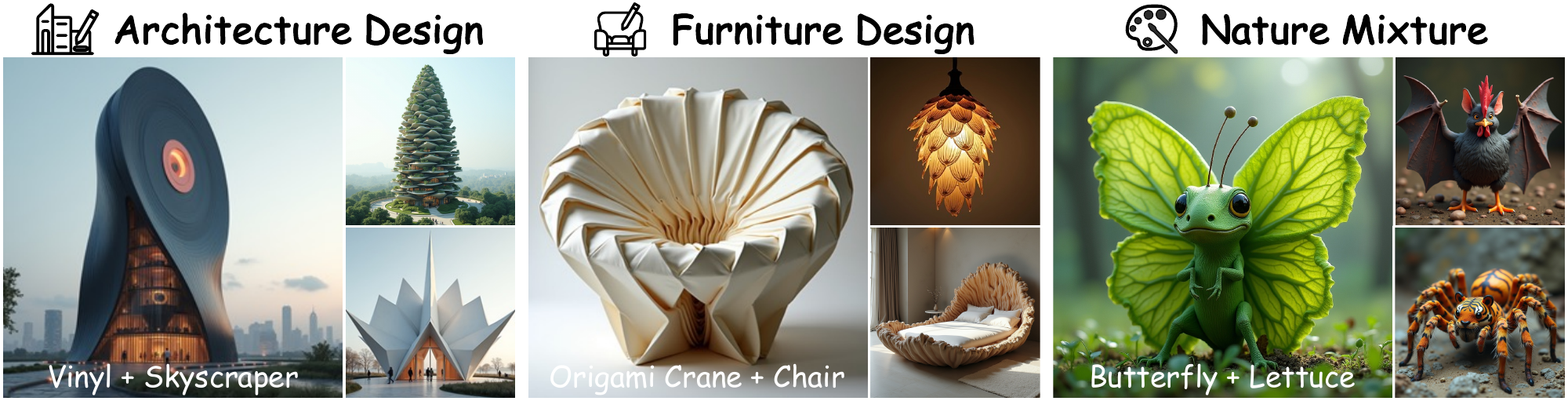}
  \vspace{-0.28in}
  \caption{\textbf{CAT: \underline{\textit{C}}reative \underline{\textit{A}}gent \underline{\textit{T}}okenization for Efficient Creative Generation.} 
  CAT generates the token template that directly injects creative intent into fuzzy prompts, enabling efficient, high-quality combinatorial creativity across \textbf{\textit{Architecture Design}}, \textbf{\textit{Furniture Design}}, and \textbf{\textit{Nature Mixture}} tasks.}
  \label{fig:teaser}
  \vspace{-0.05in}
\end{figure}

\begin{abstract}
Text-to-image (T2I) models have substantially improved image fidelity and prompt adherence, yet their creativity remains constrained by reliance on discrete natural language prompts. When presented with fuzzy prompts such as ``a creative vinyl record-inspired skyscraper'', these models often fail to infer the underlying creative intent, leaving creative ideation and prompt design largely to human users.
Recent reasoning- or agent-driven approaches iteratively augment prompts but incur high computational and monetary costs, as their instance-specific generation makes ``creativity'' costly and non-reusable, requiring repeated queries or reasoning for subsequent generations.
To address this, we introduce \textbf{CAT}, a framework for \textbf{C}reative \textbf{A}gent \textbf{T}okenization that encapsulates agents' intrinsic understanding of ``creativity'' through a \textit{Creative Tokenizer}.
Given the embeddings of fuzzy prompts, the tokenizer generates a reusable token template that can be directly concatenated with them to inject creative semantics into T2I models without repeated reasoning or prompt augmentation.
To enable this, the tokenizer is trained via creative semantic disentanglement, leveraging relations among partially overlapping concept pairs to capture the agent’s latent creative representations.
Extensive experiments on \textbf{\textit{Architecture Design}}, \textbf{\textit{Furniture Design}}, and \textbf{\textit{Nature Mixture}} tasks demonstrate that CAT provides a scalable and effective paradigm for enhancing creativity in T2I generation, achieving a $3.7\times$ speedup and a $4.8\times$ reduction in computational cost, while producing images with superior human preference and text-image alignment compared to state-of-the-art T2I models and creative generation methods.
\end{abstract}

\vspace{-0.18in}
\section{Introduction}
\vspace{-0.1in}
\begin{figure}[tb]
  \centering
  \includegraphics[width=\linewidth]{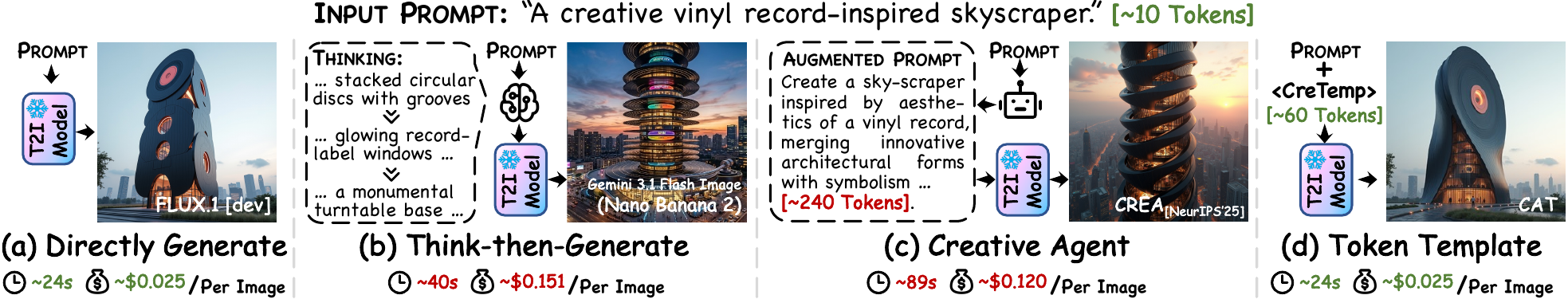}
  \vspace{-0.25in}
  \caption{\textbf{Breaking the Quality-Efficiency Bottleneck in Creative Generation.} 
  Compared with \textbf{(a)}~direct generation, \textbf{(b)}~think-then-generation and \textbf{(c)}~agent-based approaches improve conceptual fusion but incur high time and API costs (see App.~\ref{app:cost}). In contrast, \textbf{(d)}~CAT encapsulates the agent’s intrinsic understanding of ``creativity'' in a reusable token template that directly enhances semantic representations, delivering superior visual quality with minimal cost and inference time.}
  \vspace{-0.15in}
\label{fig:moti}
\end{figure}
Recent advances in text-to-image (T2I) generation have substantially improved image fidelity and controllability~\cite{bie2024renaissance, elasri2022image, wu2024multimodal}, with models such as FLUX.1~\cite{black2024} producing high-quality images that closely follow textual prompts, making prompt adherence central to precise visual control~\cite{chefer2023attend, hertzprompt}.

However, heavy reliance on prompt guidance inherently limits machine creativity~\cite{oppenlaender2022creativity, richardson2024conceptlab}, hindering T2I models from autonomously exploring novel concepts beyond human-specified semantics.
Consequently, when users provide only fuzzy prompts, e.g., ``a creative vinyl record-inspired skyscraper'', T2I models often struggle to infer the underlying creative intent and generate genuinely novel yet practically plausible images (Fig.~\ref{fig:moti}a).
In practice, creativity largely remains a human burden, requiring users to \textit{conceive novel ideas} and \textit{craft detailed prompts} to guide generation~\cite{liu2022design, hao2023optimizing}.

Existing methods for enhancing T2I models’ creative capacity primarily emphasize combinatorial generation~\cite{ha2025synthia, feng2025distribution}. BASS~\cite{li2024tp2o} and AGSwap~\cite{zhang2025agswap} perform unsupervised sampling and swapping of concept pairs, but this stochastic process is unstable and often requires repeated trials to yield genuinely creative outputs.
ConceptLab~\cite{richardson2024conceptlab} and CreTok~\cite{feng2025redefining} exploit semantic similarity to guide exploration in unknown semantic spaces; however, the supervision remains weak, constraining creativity in complex domains.
Thus, a straightforward strategy is to augment or refine fuzzy prompts, enabling T2I models to better interpret and realize creative concepts~\cite{cao2023beautifulprompt, wang2024genartist, venkateshcrea}.

Recent methods leverage reasoning-capable (Fig.~\ref{fig:moti}b) or agent-based systems (Fig.~\ref{fig:moti}c) to infer and generate creativity from fuzzy prompts~\cite{wang2024genartist, chen2025t2i, venkateshcrea}. 
However, these approaches rely on iterative prompt augmentation, \textbf{incurring substantial computational and monetary costs}, and remain constrained by semantic boundaries, limiting their ability to generate creativity that cannot be fully described in language.
Moreover, such creative insights are transient and instance-specific; for example, a prompt derived from \textit{``a vinyl record-inspired skyscraper''} cannot be easily transferred to \textit{``a mushroom-inspired pavilion''}, requiring \textbf{repeated and costly iterations for each new generation}.

This motivates an alternative perspective: \textit{can we encapsulate an agent’s \textbf{intrinsic understanding of ``creativity''} into reusable tokens, allowing us to bypass repeated queries and directly leverage these representations for creative generation?}
To this end, we introduce \textbf{CAT}, a framework for \textbf{C}reative \textbf{A}gent \textbf{T}okenization that disentangles an agent’s understanding of creativity from its augmented prompts and encapsulates this knowledge into a reusable token template, enabling structured, controllable, and efficient generation of creative concepts.

Specifically, we iteratively sample concept pairs $(c_i, s_j)$, where $c_i$ denotes the primary concept and $s_j$ provides stylistic guidance, and construct fuzzy prompts of the form ``a creative $s_j$-inspired~$c_i$''. 
These prompts are then creatively augmented by an agent, and both the fuzzy and augmented prompts are encoded by a text encoder into fuzzy embeddings $\mathcal{F}_{(c_i,s_j)}$ and augmented embeddings $\mathcal{A}_{(c_i,s_j)}$.
To tokenize the creative agent, we introduce a \textbf{Creative Tokenizer} that generates a reusable \textit{token template} from each fuzzy embedding. 
This template can be directly concatenated with the corresponding fuzzy embedding to form the creative embedding $\mathcal{C}_{(c_i,s_j)}$, which are then fed into T2I models to guide creative generation.
The tokenizer is trained via semantic disentanglement to extract the agent’s intrinsic understanding of ``creativity''.
Specifically, we exploit relations among partially overlapping concept pairs (e.g., $(c_i,s_j)$ and $(c_i,s_k)$) and enforce cosine-distance consistency between augmented embeddings and their corresponding creative embeddings, preserving the relational geometry of creative compositions and thereby promoting disentanglement between concept semantics and intrinsic creative representations.

CAT is trained on common classes from CangJie~\cite{feng2025redefining} and extended to practical domains, such as architecture and furniture design, beyond purely animal- or plant-based combinatorial creativity, enabling more practical creative generation.
Compared with state-of-the-art agent-based creative generation methods, CAT achieves substantial efficiency gains, offering a 3.7$\times$ speedup over T2I-Copilot and CREA (24s vs. 89s) and a 4.8$\times$ reduction in computational cost (\$0.025 vs. \$0.120) by eliminating redundant reasoning and agent queries. 
In addition to efficiency gains, CAT-generated images surpass diffusion-based creative generation methods in both human preference and text-image alignment. 
Evaluations with GPT-4o and user study further confirm CAT’s superiority in concept integration and originality, demonstrating its effectiveness for creative image generation.

Our contributions are summarized as follows: 
(1)~We present the first attempt to tokenize agents by encapsulating their specific knowledge into reusable tokens, eliminating repeated reasoning and agent queries to improve generation efficiency.
(2)~We propose CAT, which trains the first Creative Tokenizer by disentangling agents’ intrinsic creative knowledge, enabling efficient creative generation that surpasses the capabilities of natural language descriptions.
(3)~We extend the CangJie dataset to practical applications such as architectural and furniture design. 
Experiments validate the effectiveness and generality of our approach, particularly for real-world creative generation tasks.

\vspace{-0.06in}
\section{Related Work}
\vspace{-0.05in}
\textbf{Creative Generation.}
Recent T2I models can generate semantically coherent images from text prompts~\cite{ramesh2022hierarchical, rombach2022high, saharia2022photorealistic}, yet producing genuinely novel concepts with unexpected aesthetic or functional properties remains challenging~\cite{epstein2023art, somepalli2023diffusion}, making creative generation a key direction for advancing machine intelligence~\cite{franceschelli2024creativity, richardson2024conceptlab}.
Beyond image-level editing and fusion~\cite{avrahami2023blended, brooks2023instructpix2pix, meng2021sdedit}, recent approaches advance T2I creativity at the text level~\cite{hao2023optimizing, lian2024llmgrounded}. 
Specifically, BASS~\cite{li2024tp2o} and AGSwap~\cite{zhang2025agswap} generate novel concept combinations by sampling and swapping concept embeddings, whereas CreTok~\cite{feng2025redefining} encodes combinatorial semantics through a dedicated token.
These methods, including others~\cite{golan2025vlm, han2025enhancing, li2025rmler}, often lack explicit supervision, rely on repeated trial-and-error to uncover genuinely creative concepts, and have limited ability to integrate disparate domains, such as architectural design.
Thus, we introduce CAT, which uses a creative agent as weak supervision to disentangle abstract creative semantics from generated outputs into a token template, enabling efficient creative generation.

\textbf{Agent for Image Generation.}
LLM- and VLM-based agents have been integrated into T2I synthesis to better align human intent with model outputs~\cite{shen2023hugginggpt, wu2023visual}.
Recent works employ agents in image generation~\cite{cao2023beautifulprompt, chen2025t2i, lian2024llmgrounded, wang2024genartist, yang2024mastering}, either as automated prompt engineers that enrich sparse inputs~\cite{cao2023beautifulprompt, wang2024genartist} or as interactive assistants and compositional planners guiding generation via multi-turn feedback or layout reasoning~\cite{chen2025t2i, lian2024llmgrounded, yang2024mastering}.
Building on these capabilities, agents have been applied to creative generation, expanding prompts with detailed creative descriptions, as in CREA~\cite{venkateshcrea}, which leverages structured compositional reasoning to guide imaginative design.
However, existing agent-driven frameworks are limited by their reliance on natural language, which cannot convey fine-grained creative intent. 
We address this by tokenizing the agent into a reusable, high-dimensional token template, enabling precise and efficient creative generation with no need for repeated agent queries.

\vspace{-0.06in}
\section{Methods}
\vspace{-0.05in}
\label{sec:methods}
\textbf{Creative Agent Tokenization (CAT)} encapsulates an agent's intrinsic understanding of ``creativity'' into a reusable token template for efficient creative generation.  
We first present \textbf{Problem Formulation} (Sec.~\ref{sec:preliminaries}) and \textbf{Creative Agent System} (Sec.~\ref{sec:agent}), which augments and filters fuzzy prompts to provide weak creative supervision.  
We then introduce the \textbf{Token Template} (Sec.~\ref{sec:tokenizer}), generated by the Creative Tokenizer from fuzzy embeddings, and describe \textbf{Creative Semantic Disentanglement} (Sec.~\ref{sec:disen}), which separates abstract creative semantics from agent outputs to train the tokenizer.

\vspace{-0.06in}
\subsection{Problem Formulation of Creative Generation}
\vspace{-0.04in}
\label{sec:preliminaries}
The pursuit of machine intelligence has motivated research on creative text-to-image generation. %~\cite{xx, xx}
Creative Text-to-Image (CT2I) task~\cite{richardson2024conceptlab} aims to produce  novel visual concepts beyond conventional descriptions, while
Creative Text Pair-to-Object (TP2O) task~\cite{li2024tp2o} generates novel objects by combining distinct text concepts into coherent compositions, offering greater controllability and practical relevance, and has consequently received broader attention~\cite{feng2025distribution, feng2025redefining,  ha2025synthia, li2025rmler, zhang2025agswap}.

Building on this combinatorial generation paradigm, we formulate creative generation as a \textbf{Concept-Biased Semantic Fusion} task, capturing the inherent asymmetry in fuzzy prompts such as ``a creative $s$-inspired $c$,'' where $c$ and $s$ are two distinct concepts representing the primary concept and the style, respectively.
Conditioned on such a fuzzy prompt, standard T2I models often fail to coherently fuse the concepts (see Fig.~\ref{fig:moti}a and Fig.~\ref{fig:archi}). 
We therefore aim to learn a token template that, when concatenated with the fuzzy prompt, guides the base T2I model to synthesize novel concepts that harmoniously and creatively combine the attributes of both concepts.

\begin{figure}[tb]
  \centering
  \includegraphics[width=\linewidth]{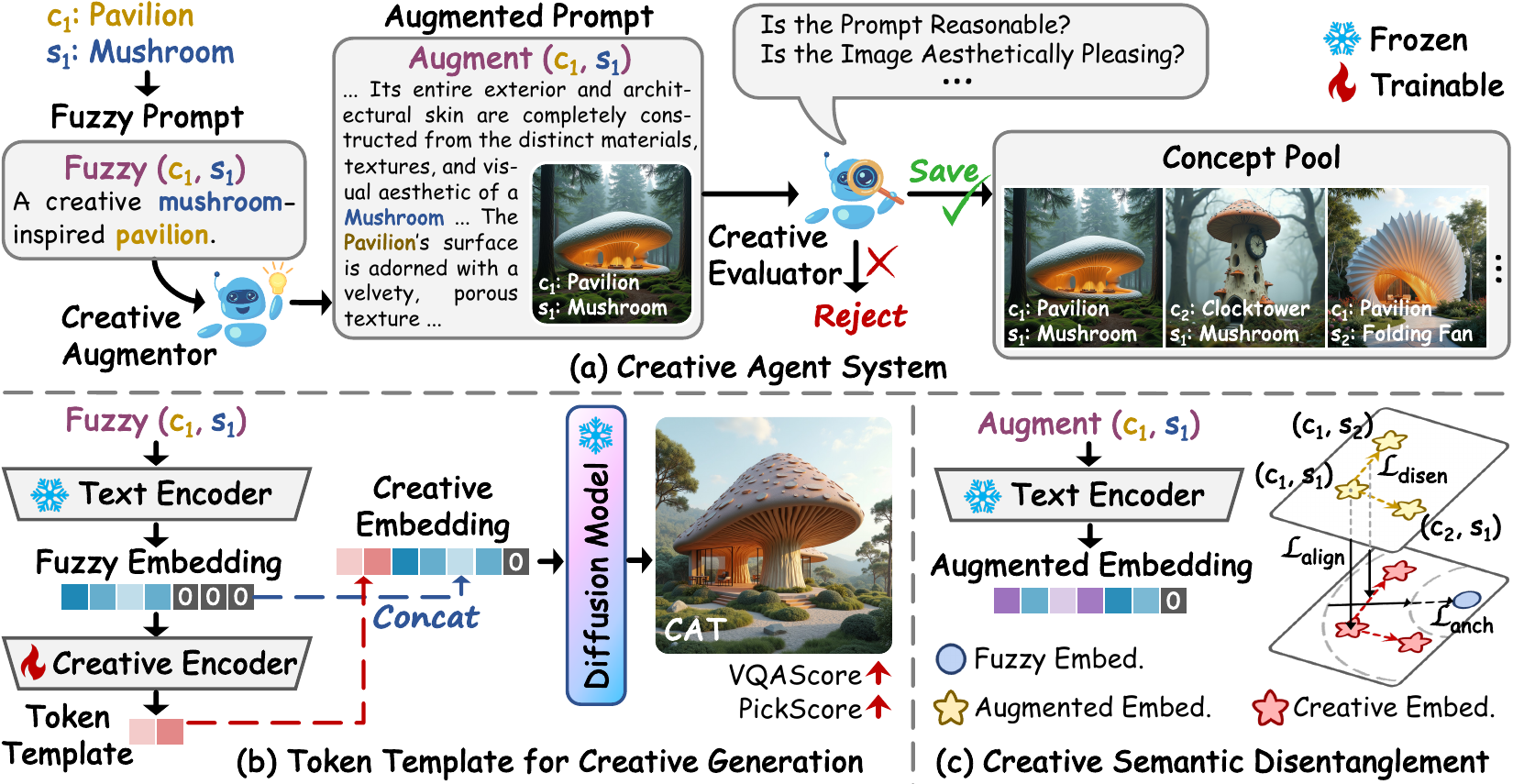}
  \vspace{-0.27in}
  \caption{\textbf{Overview of CAT.}
   % CAT aims to encapsulate the agent’s intrinsic understanding of ``creativity'' into a token template for creative image generation. 
   \textbf{(a)}~We introduce a \textit{Creative Augmentor} to augment fuzzy prompts and a \textit{Creative Evaluator} to filter generated concepts, with valid ones stored in a \textit{Concept Pool}.
   \textbf{(b)}~The \textit{Creative Tokenizer} maps each fuzzy prompt embedding to a corresponding token template, which is directly concatenated with the fuzzy embedding to form a creative embedding for creative generation.
   \textbf{(c)}~The \textit{Creative Tokenizer} is trained via semantic disentanglement, leveraging relations among partially overlapping concept pairs to capture the agent’s implicit understanding of creativity.}
  \vspace{-0.15in}
\end{figure}

\vspace{-0.06in}
\subsection{Creative Agent System for Weak Creative Supervision}
\vspace{-0.04in}
\label{sec:agent}
Recent advances in text-to-image synthesis often leverage VLM-agent systems as auxiliary tools during inference, performing tasks such as prompt refinement or spatial layout generation~\cite{chen2025t2i, venkateshcrea, wang2024genartist}.
In CAT, agents serve solely as weak supervision for creative generation, providing concept-level understanding during interaction to enable disentanglement of intrinsic creative knowledge~(Sec.~\ref{sec:tokenizer}). 
Unlike fully agent-based methods such as CREA~\cite{venkateshcrea}, which rely on iterative generate-evaluate-refine loops, CAT uses only two lightweight agents—a \textbf{Creative Augmentor} and a \textbf{Creative Evaluator}—for simple offline prompt augmentation.

Specifically, in each iteration we randomly sample two concepts—a primary concept $c_i$ and a style guide $s_j$—and construct a fuzzy prompt in the concept-biased format (i.e., ``\texttt{Fuzzy}($c_i$, $s_j$):~A creative $s_j$-inspired $c_i$'', as in Sec.~\ref{sec:preliminaries}).
The Creative Augmentor expands each fuzzy prompt \texttt{Fuzzy}($\cdot, \cdot$) into an augmented prompt \texttt{Augment}($\cdot, \cdot$) according to its \textit{intrinsic understanding} of ``creativity'', integrating $s_j$ coherently and implicitly while preserving the integrity of the primary concept $c_i$.

Although the Creative Augmentor generates detailed prompts, textual guidance alone cannot ensure successful image synthesis due to unsuitable concept combinations. 
We therefore filter the generated \texttt{Augment}($\cdot, \cdot$) prompts and corresponding images using the Creative Evaluator, which assigns a binary score $v_{ij} \in \{0,1\}$ for each concept pair $(c_i, s_j)$, with $v_{ij}=0$ indicating failed fusion, including side-by-side composition, loss of the primary concept, or absence of the target style.
The validated prompts are stored in a \textit{Concept Pool} for subsequent creative disentanglement and agent tokenization.

\vspace{-0.02in}
\subsection{Token Template for Efficient Creative Generation}
\vspace{-0.03in}
\label{sec:tokenizer}
As discussed earlier, existing agent-based approaches (e.g., CREA~\cite{venkateshcrea}) require repeated agent queries for each concept pair, resulting in significant computational overhead. 
Moreover, as agent outputs are expressed in discrete natural language, such methods remain constrained by linguistic expressibility and may fail to capture creative concepts \textit{beyond what language can explicitly describe}.

Thus, we introduce the token template, a reusable set of tokens that encapsulates the agent’s intrinsic understanding of ``creativity''.
When concatenated with fuzzy prompts, they provide the base T2I model with direct access to abstract creative semantics, enabling efficient creative generation without repeated agent queries for each concept pair.
To enhance the adaptability and concept specificity of the token template and increase the diversity of creative semantics, we introduce a Creative Tokenizer, a general generator that produces specific token template conditioned on each concept pair, rather than relying on a globally shared template as in CreTok~\cite{feng2025redefining}.

Specifically, creative generation using the token template proceeds as follows.
Given a concept pair $(c_i, s_j)$, we first encode the corresponding fuzzy prompt $\texttt{Fuzzy}(c_i, s_j)$ with a pre-trained text encoder (e.g., T5~\cite{raffel2020exploring}) to obtain fuzzy embeddings $\mathcal{F}_{(c_i, s_j)}$.
These embeddings are processed by the Creative Tokenizer to produce concept-pair-specific token template $\mathcal{T}_{(c_i, s_j)}$, which are then concatenated with $\mathcal{F}_{(c_i, s_j)}$ to form the final creative embedding $\mathcal{C}_{(c_i, s_j)}$. 
The creative embedding $\mathcal{C}_{(c_i, s_j)}$ can be directly input to a base T2I model (e.g., FLUX) to synthesize images conditioned on $(c_i, s_j)$.

\subsection{Creative Semantic Disentanglement for Training the Tokenizer}
\label{sec:disen}
Building on the token template and Creative Tokenizer introduced in Sec.~\ref{sec:tokenizer}, we now disentangle the agent's intrinsic understanding of ``creativity'' from outputs stored in the concept pool and use this knowledge to train the Creative Tokenizer, enabling it to encode the agent’s creative reasoning into reusable token representations.

\textbf{Basic Semantic Alignment.} To ensure that the generated creative embeddings preserve the agent’s core creative semantics, we first align them with the embeddings of the corresponding augmented prompts.  
Formally, let the creative embedding be $\mathcal{C}_{(c_i, s_j)} \in \mathbb{R}^{L_\mathcal{C} \times d}$ and the augmented embedding, obtained by encoding $\texttt{Augment}(c_i, s_j)$ with a pre-trained text encoder, be $\mathcal{A}_{(c_i, s_j)} \in \mathbb{R}^{L_\mathcal{A} \times d}$, where $L_\mathcal{C}$ and $L_\mathcal{A}$ denote sequence lengths and $d$ the feature dimension.
Since $L_\mathcal{C} \ll L_\mathcal{A}$, we align the global feature distributions of $\mathcal{C}_{(c_i, s_j)}$ and $\mathcal{A}_{(c_i, s_j)}$ along the sequence dimension by matching their mean and standard deviation:
\begin{equation}
    \mathcal{L}_{\text{sim}} = \|\mu(\mathcal{C}_{(c_i, s_j)}) - \mu(\mathcal{A}_{(c_i, s_j)})\|_2^2 + \|\sigma(\mathcal{C}_{(c_i, s_j)}) - \sigma(\mathcal{A}_{(c_i, s_j)})\|_2^2,
\end{equation}
where $\mu(\cdot), \sigma(\cdot) \in \mathbb{R}^d$ are computed over the sequence dimension.
To mitigate numerical instability, we apply an $L_2$ regularization on the generated token template $\mathcal{T}_{(c_i, s_j)}$:
\begin{equation}
    \mathcal{L}_{\text{reg}} = \frac{1}{L_\mathcal{C}} \sum_{t=1}^{L_\mathcal{C}} \|\mathcal{T}_{(c_i, s_j)}[t]\|_2^2,
\end{equation}
The overall loss for basic semantic alignment is then defined as $\mathcal{L}_{\text{align}} = \mathcal{L}_{\text{sim}} + \mathcal{L}_{\text{reg}}$.

\textbf{Relative Creative Disentanglement.} 
To preserve the agent's intrinsic understanding of ``creativity'', we leverage the relative relationships among overlapping concept pairs to disentangle concept- and style-specific attributes and guide learning of compositional transitions in the embedding space.  

Specifically, for each sample, we construct a local semantic triplet of partially overlapping concept pairs, 
$\mathcal{P} = \{(c_i, s_j), (c_i, s_k), (c_m, s_j)\}$, extracted from the Concept Pool. 
Within this triplet, $(c_i, s_j)$ serves as a shared anchor, inducing variations along both the concept dimensions (i.e., $(c_i, s_j) \rightarrow (c_m, s_j)$) and style dimensions (i.e., $(c_i, s_j) \rightarrow (c_i, s_k)$).
Rather than modeling absolute representations, we focus on \textbf{\textit{preserving relative semantic transitions within the triplet}}.

For any two pairs in $\mathcal{P}$, e.g., $(c_i, s_j)$ and $(c_m, s_j)$, we define their transition vectors in the creative and augmented embedding spaces as
$\Delta_{\mathcal{C}}^{(i,j \rightarrow m,j)} = \mu(\mathcal{C}_{(c_m, s_j)}) - \mu(\mathcal{C}_{(c_i, s_j)})$ and 
$\Delta_{\mathcal{A}}^{(i,j \rightarrow m,j)} = \mu(\mathcal{A}_{(c_m, s_j)})  - \mu(\mathcal{A}_{(c_i, s_j)})$, respectively. 
These vectors (i.e., $\Delta_{\mathcal{C}}^{(i,j \rightarrow m,j)}$ and $\Delta_{\mathcal{A}}^{(i,j \rightarrow m,j)}$) capture the relative semantic variations between primary concepts.
Notably, transition vectors $\Delta^{(i,j \rightarrow i,k)}$ correspond to variations in style while keeping the primary concept fixed, whereas transition vectors $\Delta^{(i,k \rightarrow m,j)}$ capture joint variations across both primary concept and style dimensions.

The relative disentanglement loss aligns the transition vectors across all pairwise edges in the triplet:
\begin{equation}
    \mathcal{L}_{\text{disen}} = \frac{1}{|\mathcal{E}|} \sum_{e \in \mathcal{E}} \Big( 1 - \cos\big(\Delta_{\mathcal{C}}^{e}, \Delta_{\mathcal{A}}^{e}\big) \Big), \quad \mathcal{E} = \Big\{ \underbrace{(i,j \rightarrow i,k)}_{\text{Style}}, \underbrace{(i,j \rightarrow m,j)}_{\text{Concept}}, \underbrace{(i,k \rightarrow m,j)}_{\text{Joint}} \Big\}.
\end{equation}
By explicitly minimizing the cosine distance across these corresponding edges, the model synchronizes the two latent topologies of the creative and augmented embedding spaces.
This alignment preserves the directional semantic variations within the triplet while maintaining its relative geometric structure, thereby explicitly disentangling concept and style attributes and isolating the underlying creative factors.

\textbf{Elastic Semantic Anchoring.}
While disentangling concept and style helps isolate the intrinsic semantics of creativity, excessive disentanglement may induce semantic drift from the base concept. 
To prevent this, we introduce an elastic semantic anchor implemented via a margin-based hinge loss. 
For a concept pair $(c_i, s_j)$, the creative embedding $\mathcal{C}_{(c_i, s_j)}$ is constrained to remain close to its corresponding fuzzy embedding $\mathcal{F}_{(c_i, s_j)}$:
\begin{equation}
    \mathcal{L}_{\text{anch}} =
    \max\!\left(0,\; m - \cos\big(\mu(\mathcal{C}_{(c_i, s_j)}),\; \mu(\mathcal{F}_{(c_i, s_j)})\big)\right).
\end{equation}
Here, $m$ is a predefined margin that penalizes excessive deviation from the base concept.

\textbf{Total Objective.} The overall training objective for the creative tokenizer is formulated as:
\begin{equation}
    \mathcal{L}_{\text{total}} = \mathcal{L}_{\text{align}} + \alpha \mathcal{L}_{\text{disen}} + \beta\mathcal{L}_{\text{anch}}
\label{eq:loss}
\end{equation}
where $\alpha$ and $\beta$ are hyperparameters that balance the contributions of each loss.

\vspace{-0.05in}
\section{Experiments}
\vspace{-0.05in}
\textbf{Datasets.}
We adopt the \textit{CangJie} dataset~\cite{feng2025redefining}, which comprises 60 base concepts spanning animals and plants.  
To support practical creative design, we extend it beyond \textit{\textbf{Nature Mixture}} to \textit{\textbf{Architecture Design}} (10 primary and 55 style-guiding concepts) and \textit{\textbf{Furniture Design}} (20 primary and 38 style-guiding concepts) (see App.~\ref{sec:data details} for details).

\textbf{Experimental Setup.}
We use FLUX.1~\cite{black2024} as the base model with a dual-stream text encoder (CLIP-L/14~\cite{radford2021learning} and T5-XXL~\cite{raffel2020exploring}).  
To validate architectural generalizability, we also evaluate CAT on Stable Diffusion 3~\cite{esser2024scaling}.
The Creative Tokenizer is a lightweight 2-layer Transformer that encodes each fuzzy embeddings to generate a corresponding token template of length 64 in the T5 space, along with a linear projection that produces a global feature vector in the CLIP space (see App.~\ref{sec:ct details} for details).
The Creative Tokenizer is trained for 500 epochs using AdamW with learning rate $10^{-4}$, using balance weights $\alpha = 1.5$, $\beta = 0.1$, and anchor margin $m = 0.85$.

\textbf{Evaluation Metrics.}
We evaluate performance with VQAScore~\cite{li2024tp2o}, PickScore~\cite{kirstain2023pick}, and ImageReward~\cite{xu2024imagereward} for prompt alignment, aesthetics, and human preference. 
Creativity is further assessed via GPT-4o~\cite{hurst2024gpt} and a user study, focusing on conceptual integration and originality (see App.~\ref{sec:eval details}).

\vspace{-0.05in}
\section{Results}
\vspace{-0.05in}
\label{sec:results}
\begin{figure}[tb]
  \centering
  \includegraphics[width=\linewidth]{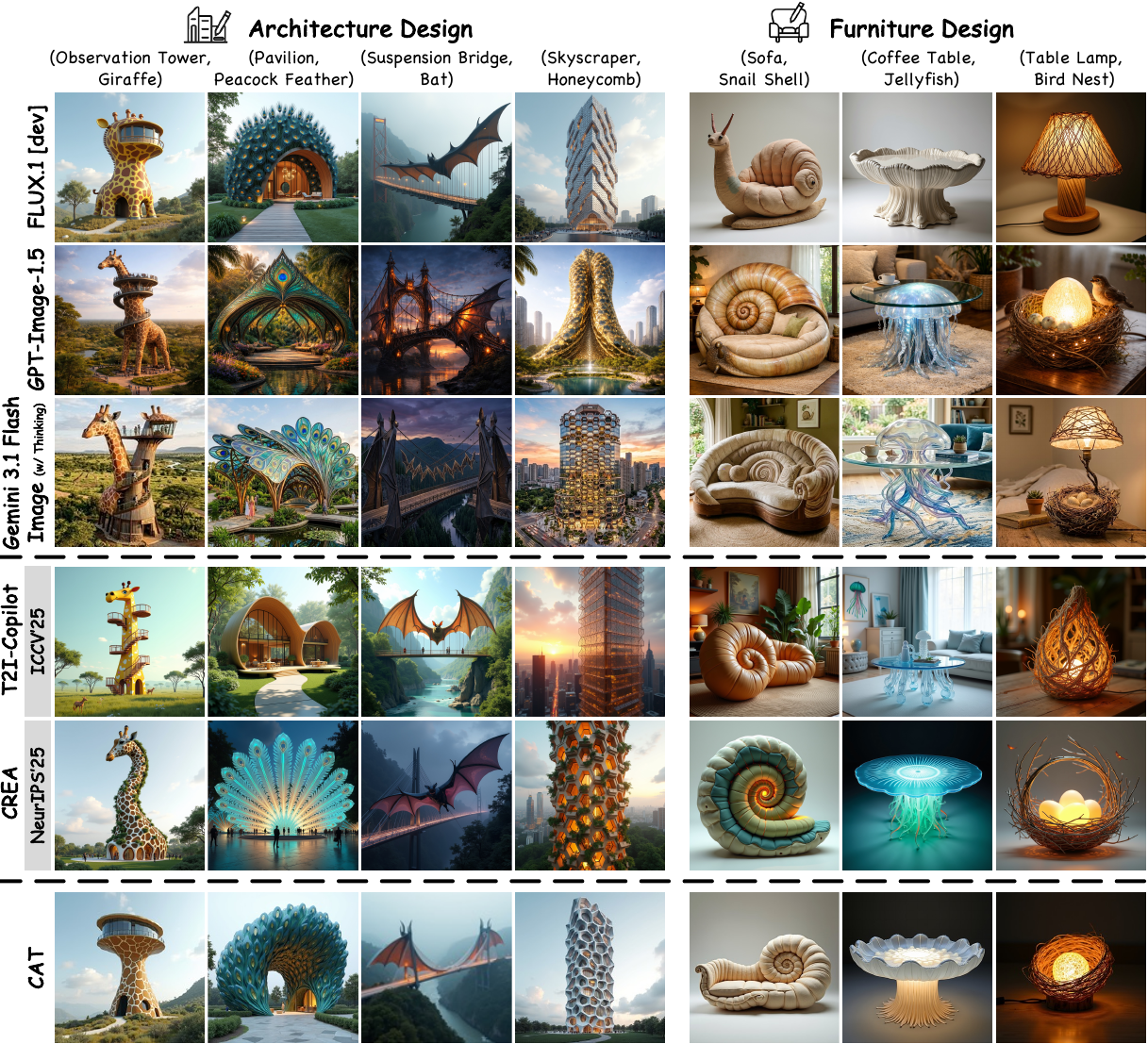}
  \vspace{-0.25in}
  \caption{\textbf{Performance of CAT on the \textit{Architecture Design} and \textit{Furniture Design} Tasks}.
  We compare CAT with the representative open-source T2I model FLUX.1~\cite{black2024}, proprietary models GPT-Image-1.5~\cite{openai2025gptimage15} and Gemini 3.1 Flash Image~\cite{google2026geminiflashimage} (enhanced with Gemini-3.1-Pro for complex reasoning), as well as agent-driven creative generation methods T2I-Copilot~\cite{chen2025t2i} and CREA~\cite{venkateshcrea}.}
  \vspace{-0.1in}
\label{fig:archi}
\end{figure}

\begin{figure}[tb]
  \centering
  \includegraphics[width=\linewidth]{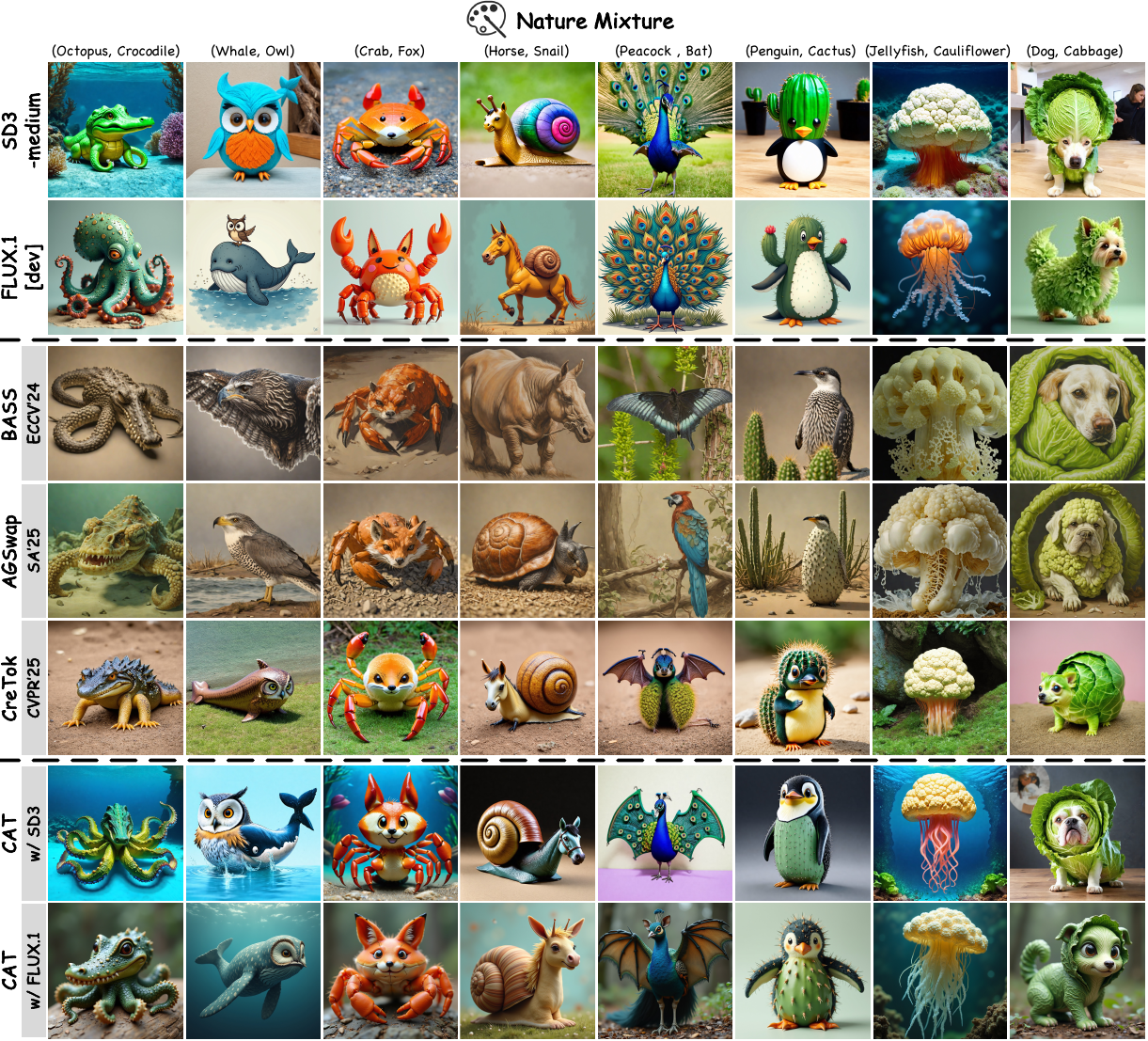}
  \vspace{-0.25in}
  \caption{\textbf{Performance of CAT on the \textit{Nature Mixture} Task.} 
  We compare CAT with representative T2I models, including FLUX.1~\cite{black2024} and Stable Diffusion 3 (SD3)~\cite{esser2024scaling}, as well as state-of-the-art creative generation methods such as BASS~\cite{li2024tp2o}, AGSwap~\cite{zhang2025agswap}, and CreTok~\cite{feng2025redefining}.}
  \vspace{-0.1in}
\label{fig:animal}
\end{figure}

\subsection{Performance on Creative Architecture \& Furniture Design}
\label{sec:design}
We first assess CAT’s capability for practical creative design in \textit{\textbf{Architecture Design}} and \textit{\textbf{Furniture Design}}, with representative results shown in Fig.~\ref{fig:archi}.
Despite the strengths of existing baselines, they struggle to generate conceptually blended artifacts in domains requiring both aesthetic form and functional utility. 
Our qualitative analysis identifies three primary failure modes:
\begin{itemize}[leftmargin=20pt,itemsep=0pt]
    \item \textbf{\textit{Stylistic Degradation:}} 
    Models fail to accurately render the intended material textures or aesthetic attributes on the structure, often reverting to default priors, as in \texttt{(Skyscraper, Honeycomb)} by FLUX.1 and \texttt{(Pavilion, Peacock Feather)} by T2I-Copilot.
    \item \textbf{\textit{Concept Deprivation:}} 
    The base object loses its functional and structural integrity, collapsing into a biological texture devoid of architectural logic, as in \texttt{(Skyscraper, Honeycomb)} by GPT-Image-1.5 and \texttt{(Pavilion, Peacock Feather)} by CREA.
    \item \textbf{\textit{Superficial Composition:}} 
    Models perform shallow spatial merging rather than coherent structural integration, as in \texttt{(Observation Tower, Giraffe)} across all baselines.
\end{itemize}

Despite extensive pre-training, the open-source FLUX model exhibits pronounced \textit{Concept Deprivation} and \textit{Superficial Composition}, failing to achieve coherent creative designs. 
Proprietary models, although benefiting from larger datasets and LLM-based reasoning, still produce superficial style-base combinations without deeper semantic integration.
Existing agent-based methods aim to enhance creativity via detailed prompting; however, this often causes overfitting to redundant local features and semantic drift, leading to the erosion of core concepts and triggering \textit{Stylistic Degradation}.

CAT bypasses discrete natural language prompts that induce semantic drift, directly encoding the agent’s intrinsic understanding of creativity into a continuous, high-dimensional token template.
Operating in this dense latent space avoids the limitations of discrete language and prevents over-attention to redundant text.
Consequently, CAT achieves deep integration of diverse concepts, eliminating \textit{Superficial Composition}, while preserving the base object’s functional semantics and seamlessly applying intended aesthetic priors, thus addressing both \textit{Concept Deprivation} and \textit{Stylistic Degradation}.
Moreover, the token template allows CAT to generate creative images without costly LLM reasoning or agent queries, adding negligible overhead compared to direct generation (Fig.~\ref{fig:moti}).

\vspace{-0.1in}
\subsection{Performance on Creative Nature Mixture}
\vspace{-0.1in}
Unlike the Creative Design tasks in Sec.~\ref{sec:design}, Creative Mixture tasks demand the synthesis of concepts that have never appeared before, as illustrated in Fig.~\ref{fig:animal}.
Our qualitative analysis shows that foundational T2I models themselves (e.g., FLUX.1) still struggle to effectively blend novel concepts, suffering from \textit{Concept Deprivation} and \textit{Superficial Composition}.
While methods tailored for creative natural mixture tasks (e.g., AGSwap~\cite{zhang2025agswap} and CreTok~\cite{feng2025redefining}) partially address these alignment failures, their rigid integration strategies still compromise coherence and overall aesthetic quality.

In contrast, by embedding diverse creative intents directly into the reusable token template, CAT facilitates deep semantic integration, surpassing the coarse blending of existing methods and generating highly creative mixtures while preserving both aesthetic fidelity and conceptual coherence.
To evaluate robustness across T2I models, we conduct experiments on two mainstream diffusion models, FLUX.1 and Stable Diffusion 3. Despite their distinct multi-text encoder architectures and stylistic priors, CAT consistently achieves robust, high-quality results.
Moreover, even on the same T2I backbone (i.e., Stable Diffusion 3), CAT significantly outperforms state-of-the-art creative generation methods (e.g., AGSawp and CreTok), producing images with higher aesthetic quality (e.g., \texttt{(Dog, Cabbage)}) and more coherent conceptual fusion (e.g., \texttt{(Jellyfish, Cauliflower)}).

\vspace{-0.05in}
\subsection{Evaluation for Creativity}
\vspace{-0.05in}
\textbf{Quantitative Comparisons.}
We evaluate generation quality using VQAScore, PickScore, and ImageReward, with results in Table~\ref{tab:eval}a showing that CAT consistently outperforms all baseline methods across these metrics.
This indicates that CAT achieves stronger alignment with human aesthetics and demonstrates improved prompt alignment and visual coherence, resulting in more semantically consistent and visually compelling concept fusion.
Given the limitations of existing metrics in capturing out-of-distribution creativity, we further employ GPT-4o for evaluation (see prompts in App.~\ref{subsec:gpt prompt}). 
Results in Table~\ref{tab:eval}b (i.e., VLM-as-a-judge) show clear improvements in originality and aesthetics (see Table~\ref{tab:gpt4o_eval} in App.~\ref{subsec:gpt prompt}) across both creative design and mixture tasks, highlighting CAT’s ability to synthesize entirely novel creative concepts.

\begin{table}[t]
    \centering
    \setlength{\tabcolsep}{0.6 mm} 
    \caption{\textbf{Quantitative Comparison of Creative Generation.} Performance is measured using automated image-text and human-aligned metrics, with creativity further evaluated by GPT-4o, and complemented by a user study reporting average human preference ranks.}
    \vspace{-0.07in}
    \resizebox{\textwidth}{!}{
    \begin{tabular}{@{}ll ccc ccc cccc@{}} 
    \toprule[1.3pt]
     & & \multicolumn{3}{c}{\textbf{Architecture Design}} & \multicolumn{3}{c}{\textbf{Furniture Design}} & \multicolumn{4}{c}{\textbf{Nature Mixture}} \\
     \cmidrule(lr){3-5} 
     \cmidrule(lr){6-8} 
     \cmidrule(l){9-12} 
     & & T2I-Copilot & CREA & \cellcolor{blue!12}{CAT} 
     & T2I-Copilot & CREA & \cellcolor{blue!12}{CAT}  
     & BASS & AGSwap & CreTok & \cellcolor{blue!12}{CAT} \\
     \midrule[1.3pt]
     \multirow{3}{*}{\textbf{(a)}} & VQAScore$\uparrow$ 
     & 0.80 & 0.80 & \cellcolor{blue!12}{\textbf{0.85}} 
     & 0.87 & 0.88 & \cellcolor{blue!12}{\textbf{0.91}} 
     & 0.76 & 0.76 & 0.85 & \cellcolor{blue!12}{\textbf{0.89}} \\
     & PickScore$\uparrow$ 
     & 23.51 & 23.64 & \cellcolor{blue!12}{\textbf{24.11}} 
     & 23.29 & 22.90 & \cellcolor{blue!12}{\textbf{23.51}}
     & 21.64 & 21.90 & 22.25 & \cellcolor{blue!12}{\textbf{24.03}} \\
     & ImageReward$\uparrow$ 
     & 1.19 & 0.90 & \cellcolor{blue!12}{\textbf{1.46}} 
     & 1.14 & 0.77 & \cellcolor{blue!12}{\textbf{1.26}} 
     & 0.70 & 0.50 & 1.10 & \cellcolor{blue!12}{\textbf{1.77}} \\
     \midrule
     \textbf{(b)} & VLM-as-a-judge$\uparrow$ 
     & 7.30 & 7.82 & \cellcolor{blue!12}{\textbf{8.44}} 
     & 7.88 & 7.64 & \cellcolor{blue!12}{\textbf{8.20}} 
     & 6.81 & 7.50 & 7.59 & \cellcolor{blue!12}{\textbf{8.82}} \\
     \midrule
     \textbf{(c)} & User Study (Aver. Rank)$\downarrow$ 
     & 2.33 & 2.19 & \cellcolor{blue!12}{\textbf{1.48}}
     & 2.28 & 2.27 & \cellcolor{blue!12}{\textbf{1.44}}
     & 3.16 & 2.98 & 2.52 & \cellcolor{blue!12}{\textbf{1.34}} \\
    \bottomrule[1.3pt]
    \end{tabular}
    }
    \vspace{-0.23in}
\label{tab:eval}
\end{table}

\textbf{User Study.}
To assess human preferences in creative generation, we conduct a user study with 30 participants from diverse backgrounds (see App.~\ref{sec:app_user} for details).
As shown in Table~\ref{tab:eval}c, CAT consistently outperforms all baselines, achieving the highest average ranks of 1.48, 1.44, and 1.34 in \textbf{\textit{Architecture Design}}, \textbf{\textit{Furniture Design}}, and \textbf{\textit{Nature Mixture}}, respectively.
These results further indicate that CAT consistently generates creativity that are both aesthetically preferred and exhibit superior conceptual coherence and creative synthesis, closely aligning with human judgment.

\subsection{Ablation and Analysis}
\begin{wrapfigure}{r}{0.6\textwidth}
  \centering
  \vspace{-0.2in}
  \includegraphics[width=\linewidth]{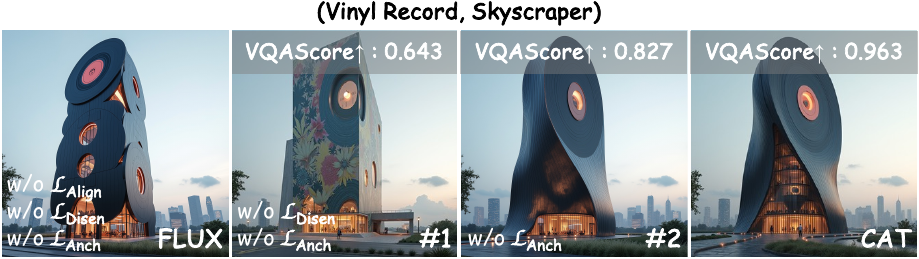}
  \vspace{-0.25in}
  \caption{\textbf{Ablation Study on the Contribution of Each Loss Component.} 
  The baseline uses direct generation with FLUX; Variant \#1 uses only $\mathcal{L}_{\text{align}}$; 
  Variant \#2 adds $\mathcal{L}_{\text{disen}}$; 
  CAT corresponds to the full model with $\mathcal{L}_{\text{total}}$.}
  \vspace{-0.1in}
  \label{fig:ablation}
\end{wrapfigure}

\textbf{Ablation of Training Loss.}
To validate the contribution of each component, we conduct a stepwise ablation study on the total loss $\mathcal{L}_{\text{total}}$ (Eq.~\ref{eq:loss}). 
As shown in Fig.~\ref{fig:ablation}, variant \#1, compared to direct generation with FLUX, exhibits more realistic architectural characteristics but shows abrupt integration.
Variant \#2 enhances concept integration by modeling relational dependencies, but may weakly preserve the original concepts, as reflected in the loss of fine-grained attributes.
In contrast, CAT achieves a balanced synthesis, preserving both aesthetic quality and functional integrity.

\begin{wrapfigure}{r}{0.6\textwidth}
  \centering
  \vspace{-0.17in}
  \includegraphics[width=\linewidth]{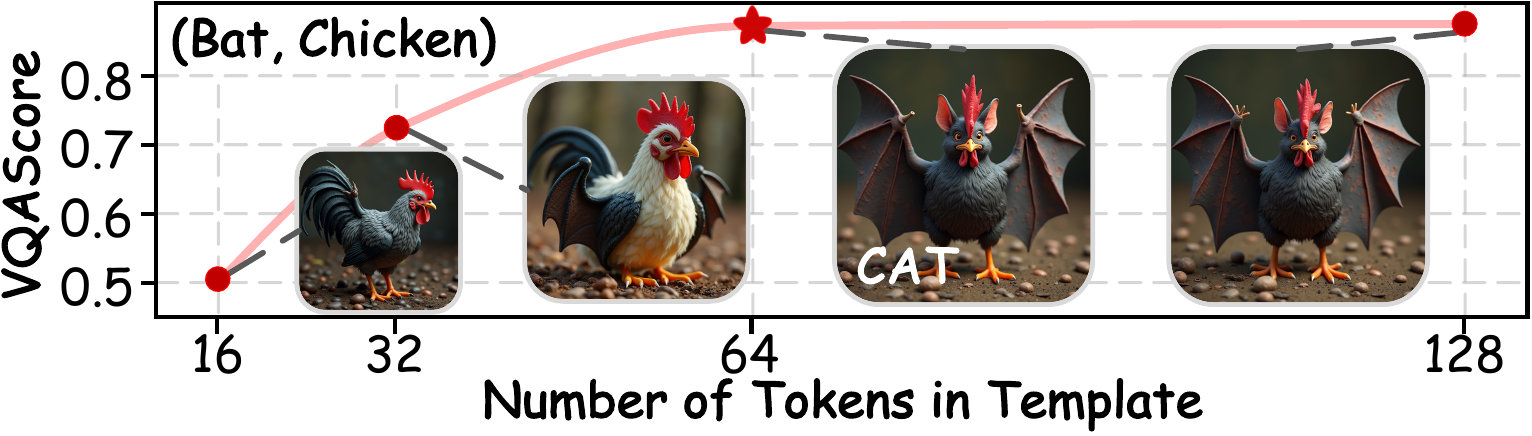}
  \vspace{-0.27in}
  \caption{\textbf{Impact of Token Number on Token Template Performance.} 
  We evaluate the performance of the token template with 16, 32, 64 (Ours), and 128 tokens.}
  \vspace{-0.1in}
  \label{fig:ana_num}
\end{wrapfigure}
\textbf{Analysis of the Number of Tokens in the Template.}
We investigate how the number of tokens in the template affects creative generation.
As shown in Fig.~\ref{fig:ana_num}, template with fewer tokens (e.g., 16 or 32) lacks sufficient capacity to encode complex creative intents, limiting their ability to capture rich semantic interactions and yielding less coherent concept integration.
Conversely, increasing the template to 128 tokens produces minimal visual improvement over the 64-token variant, but the extra capacity may inadvertently encode object-specific semantics beyond abstract creative intent, slightly reducing VQAScore.
Thus, a 64-token template offers an optimal balance, achieving smooth concept fusion without incurring extra computational overhead.

\begin{wrapfigure}{r}{0.6\textwidth}
  \centering
  \vspace{-0.2in}
  \includegraphics[width=\linewidth]{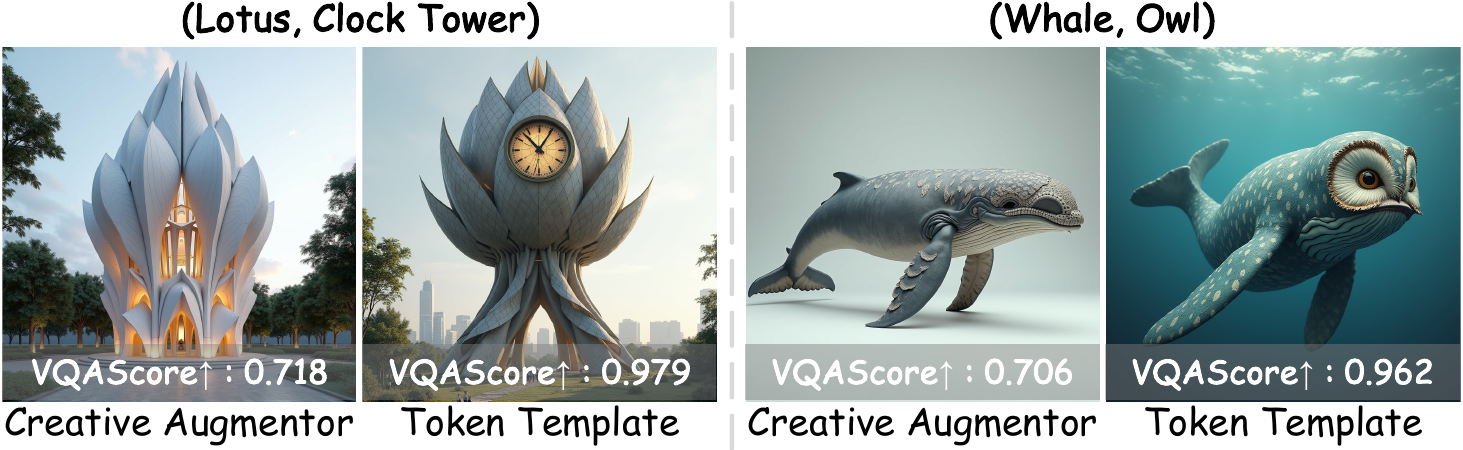}
  \vspace{-0.3in}
  \caption{\textbf{Token Template vs. Creative Augmentor.} 
  Continuous token template encodes richer creative intent, yielding more coherent and visually harmonious outputs than the Creative Augmentor using discrete language prompts.}
  \vspace{-0.1in}
  \label{fig:ana_agent}
\end{wrapfigure}

\textbf{Continuous Token Template for Creative Representation.}
Unlike discrete natural language prompts, which are limited in conveying abstract creative intent, continuous token template encodes rich, high-dimensional representations that capture structured semantic relations and abstract transformations. 
Thus, encapsulating creative intent in this dense latent space enables seamless integration of multiple concepts while preserving both aesthetic and functional integrity. 
As shown in Fig.~\ref{fig:ana_agent}, the token template consistently produces more seamless and naturally blended visuals, demonstrating their capacity to encode creative intent beyond what discrete language can express.

\vspace{-0.07in}
\section{Conclusion}
\vspace{-0.08in}
We propose CAT, a framework for efficient creative generation via Creative Agents Tokenization. CAT employs the \textit{Creative Tokenizer} to convert an agent’s intrinsic understanding of creativity into a reusable token template that directly inject creative intent into fuzzy prompts.
This approach addresses the limitations of existing agent-based and reasoning-driven methods, which are often time-consuming and constrained by the discreteness of natural language prompts, while enabling efficient, high-quality creative generation and providing a scalable framework that combines semantic flexibility with computational efficiency.
Extensive experiments show that CAT significantly enhances model creativity, outperforming state-of-the-art T2I models and creative generation methods.

% \subsubsection{prompt engineering}
% \subsection{Diversity}

% \subsubsection{style}

\section*{Acknowledgement}
We sincerely appreciate Freepik for contributing to the figure design. 
This research was supported by the Jiangsu Science Foundation (BG2024036, BK20243012), the National Natural Science Foundation of China (625B2045, 62125602, U24A20324, 92464301, 62306073), the New Cornerstone Science Foundation through the XPLORER PRIZE, the Fundamental Research Funds for the Central Universities (2242025K30024), and SEU Innovation Capability Enhancement Plan for Doctoral Students (CXJH\_SEU 26023).

\bibliography{neurips_2025}
\bibliographystyle{plain}

\clearpage
\appendix
\section{Details of Computational Cost Calculation}
\label{app:cost}
All reported execution times and the derived cost estimates in Fig.~\ref{fig:moti} are based on experiments conducted on a single NVIDIA H100 GPU.
For FLUX.1 [dev] and CAT, computational costs are estimated based on standard cloud GPU rental pricing. 
In contrast, for Gemini 3.1 Flash Image, the cost is computed directly from the official API pricing, which is \$0.151 per 4K image\footnote{\url{https://ai.google.dev/gemini-api/docs/pricing}}. 

For CREA, we limit the maximum number of iterations to 1. The cost of its agent interaction component is computed based on the actual token consumption using the official GPT-4o API token pricing\footnote{\url{https://developers.openai.com/api/docs/pricing}}, while the cost of its internal FLUX image generation process is estimated using standard GPU rental rates. 
These calculations take into account the total runtime and hardware resources required, providing a consistent and comparable measure of computational expenditure across all evaluated generation methods.

\section{Details of the Creative Agent System}
\label{sec:ca details}
In this section, we provide the comprehensive prompts utilized by the Creative Agent System, detailing their construction, intended guidance for creative generation, and role in producing augmented outputs for weak supervision.

% 新加的
In our experiments, the Creative Agent System generates 100 sets of $2 \times 2$ concept matrices for each domain. Specifically, within each generated matrix, one concept pair is randomly selected as a test sample. To prevent data leakage, these test samples are strictly excluded from the Concept Pool and do not participate in the training of the Creative Tokenizer. Ultimately, these held-out samples, together with completely unseen concept pairs not processed by the Creative Agent System, constitute our final evaluation test set.

\subsection{Prompts for the Creative Augmentor}
This subsection details the prompts used by the Creative Augmentor to generate augmented prompts from fuzzy prompts, guided by the agent’s intrinsic understanding of creativity. 

\begin{tcolorbox}[
    colback=blue!5!white,      
    colframe=blue!60!black,    
    title=\textbf{Prompts for the Creative Augmentor}, 
    arc=2mm,                   
    boxrule=0.5pt,             
    left=6pt, right=6pt, top=6pt, bottom=6pt,
    breakable 
]

\textbf{\large [Architecture \& Furniture Design]} 
\vspace{0.3em}

\texttt{[Role]} You are the Generation Agent (Teacher) specializing in Architecture and Industrial Design.

\texttt{[System Instruction]} We are providing supervision to train a decoupled AI generator. You will receive 2 Concepts and 2 Styles. Your task is to generate 4 Teacher Prompts covering every 2$\times$2 combination.

[DECOUPLING LAWS]
\begin{itemize}
    \setlength{\itemsep}{0pt}
    \setlength{\parskip}{0pt}
    \item \textbf{Concept} dictates the unbreakable physical silhouette, macro-geometry, and base function.
    \item \textbf{Style} dictates the surface material, texture, light reflection, and the surrounding atmospheric vibe.
\end{itemize}

[MANDATORY SYNTACTIC TEMPLATE] (Must be exactly 4 sentences)
\begin{enumerate}
    \setlength{\itemsep}{0pt}
    \setlength{\parskip}{0pt}
    \item (Identity \& Silhouette): ``A striking, highly detailed creative design taking the exact physical silhouette, macro-geometry, and functional form of a [Concept].''
    \item (Surface \& Material): ``Its entire exterior and architectural skin are completely constructed from the distinct materials, textures, and visual aesthetic of a [Style].''
    \item (Micro-details \& Lighting): ``[Describe 2-3 specific visual interactions...]''
    \item (Environment \& Atmosphere): ``[Describe a complementary contextual background...]''
\end{enumerate}
Constraints: Length 150-250 words. Focus on ONE main structure. Output strictly in JSON format: \texttt{\{``C1\_S1'': ``...'', ``C1\_S2'': ``...'', ...\}}

\vspace{1em} \hrule \vspace{1em} % 分割线

% ================= 分支 2 =================
\textbf{\large [Nature Mixture]} \vspace{0.3em}

\texttt{[Role]} You are the Generation Agent (Teacher) specializing in Biological \& Botanical Fusion.

\texttt{[System Instruction]} We are providing supervision to train a decoupled AI generator. You will receive 2 Concepts and 2 Styles. Your task is to generate 4 Teacher Prompts covering every 2$\times$2 combination.

[CRITICAL: THE SINGLE-ENTITY WELDING RULE] \\
You MUST generate a SINGLE, mathematically fused mutant creature. UNDER NO CIRCUMSTANCES should there be two animals standing side-by-side or interacting. The creature MUST share ONE skeleton dictated entirely by the [Concept], while its skin/covering is dictated entirely by the [Style].

[MANDATORY SYNTACTIC TEMPLATE] (Must be exactly 4 sentences)
\begin{enumerate}
    \setlength{\itemsep}{0pt}
    \setlength{\parskip}{0pt}
    \item (Base Anatomy): ``A single, unified hybrid creature that strictly inherits the exact anatomical skeleton, limb structure, and overall body posture of a [Concept].''
    \item (Epidermal Mapping): ``However, its entire epidermis is seamlessly replaced by the distinctive textures, fur, scales, or biological surface of a [Style].''
    \item (Biological Details): ``[Describe 1-2 highly specific, realistic biological details...]''
    \item (Lighting/Vibe): ``The mutant organism is rendered in photorealistic macro-photography with cinematic lighting against a clean, neutral background.''
\end{enumerate}
Output strictly in JSON format: 
\texttt{\{``C1\_S1'': ``...'', ``C1\_S2'': ``...'', ...\}}

\end{tcolorbox}

\subsection{Prompts for the Creative Evaluator}
This subsection details the prompts used by the Creative Evaluator to assess augmented prompts and their corresponding generated images in terms of alignment and visual coherence.

\begin{tcolorbox}[
    colback=green!5!white,     
    colframe=green!50!black,   
    title=\textbf{Prompts for the Creative Evaluator}, 
    arc=2mm,                   
    boxrule=0.5pt,             
    left=6pt, right=6pt, top=6pt, bottom=6pt,
    breakable
]

\textbf{\large [Text Evaluator]} 
\vspace{0.3em}

\texttt{[Role]} You are the Judgement Agent (Quality Assurance Critic).

\texttt{[System Instruction]} Your task is to review 4 text-to-image prompts generated by the Teacher Agent for a 2$\times$2 Conceptual Matrix. You must ensure they strictly follow the ``Decoupling Laws'' before we spend GPU resources rendering them.

[EVALUATION CRITERIA] (Input: JSON string of 4 prompts)
\begin{enumerate}
    \setlength{\itemsep}{0pt}
    \setlength{\parskip}{0pt}
    \item Syntax Compliance: Does EVERY prompt have exactly 4 sentences?
    \item Decoupling Fidelity: In C1\_S1 and C1\_S2, is the underlying physical structure (C1) described consistently, with ONLY the styles changing?
    \item Integration Rationality: Do the prompts describe a coherent fusion where style elements are logically integrated with the concept's structure?
\end{enumerate}

[OUTPUT FORMAT] Strict JSON:\\
\{ \\
\indent ``passed'': true, \\
\indent ``reason'': ``Brief explanation of why it passed or failed ...'' \\
\}

\vspace{1em} \hrule \vspace{1em} % 分割线

% ================= 分支 2 =================
\textbf{\large [Vision Evaluator]} \vspace{0.3em}

\texttt{[Role]} You are an expert AI Art and Design Evaluator specializing in Concept-Style Fusion.

\texttt{[System Instruction]} Your task is to inspect a 2x2 matrix of generated images demonstrating `Concept-Style Fusion'.
\begin{itemize}
    \setlength{\itemsep}{0pt}
    \setlength{\parskip}{0pt}
    \item `Concept' (C) defines the recognizable macro-structure.
    \item `Style' (S) defines the surface appearance (material/texture).
\end{itemize}

[CRITICAL EVALUATION MINDSET] \\
Be forgiving but logical. Do NOT reject images for minor artifacts, slight background clutter, or creative structural adaptations.

ONLY REJECT (passed: false) if you see these FATAL ERRORS:
\begin{enumerate}
    \setlength{\itemsep}{0pt}
    \setlength{\parskip}{0pt}
    \item The ``Side-by-Side'' Failure: Shows a normal concept and a normal style placed next to each other, instead of fusing them into ONE object.
    \item Complete Concept Erasure: The structural identity is completely destroyed and unrecognizable.
    \item Complete Style Ignorance: The object has no style texture/material applied at all.
\end{enumerate}

[OUTPUT FORMAT] Strict JSON:\\
\{ \\
\indent ``passed'': true, \\
\indent ``reason'': ``Briefly state if it fused successfully or name the fatal error.'' \\
\}
\end{tcolorbox}

\section{Details of the Creative Tokenizer}
\label{sec:ct details}
The Creative Tokenizer is designed as a lightweight module to map the input fuzzy embeddings into the distinct representation spaces of the text encoders. 

To generate the sequence of $N=64$ soft token templates for the T5 encoder, we first extract a global context $\bar{\mathbf{f}}_{(c_i, s_j)} \in \mathbb{R}^{d_{in}}$ by mean-pooling the input fuzzy embeddings $\texttt{Fuzzy}(c_i, s_j)$. This context is projected via a 2-layer Multilayer Perceptron (MLP) to predict a flattened weight vector $\mathbf{m}_{(c_i, s_j)} \in \mathbb{R}^{N \cdot d_{basis}}$:
\begin{equation}
    \mathbf{m}_{(c_i, s_j)} = \mathbf{W}_{p2}(\text{GELU}(\mathbf{W}_{p1} \bar{\mathbf{f}}_{(c_i, s_j)} + \mathbf{b}_{p1})) + \mathbf{b}_{p2}
\end{equation}
where $d_{basis} = 256$. This vector is subsequently reshaped into specific mixing weights $\mathbf{M}_{(c_i, s_j)} \in \mathbb{R}^{N \times d_{basis}}$ and multiplied by a globally shared, learnable basis dictionary $\mathbf{B} \in \mathbb{R}^{d_{basis} \times d_{in}}$ to yield the initial dynamic prefix $\mathbf{P}_{dyn}^{(c_i, s_j)} = \mathbf{M}_{(c_i, s_j)} \mathbf{B} \in \mathbb{R}^{N \times d_{in}}$.

The core of the Creative Tokenizer lies in its contextual interaction mechanism driven by a Transformer Encoder. To imbue the initial prefix with fine-grained semantic details, we concatenate $\mathbf{P}_{dyn}^{(c_i, s_j)}$ with the original unpooled $\texttt{Fuzzy}(c_i, s_j)$ along the sequence dimension. This combined sequence is then fed into a lightweight 2-layer Transformer Encoder, configured with 8 attention heads and a Pre-LayerNorm architecture, to produce the contextually enriched output $\mathbf{C}'_{(c_i, s_j)} \in \mathbb{R}^{(N + L) \times d_{in}}$:
\begin{equation}
    \mathbf{C}'_{(c_i, s_j)} = \text{TransformerEncoder}\Big(\big[\mathbf{P}_{dyn}^{(c_i, s_j)} \,;\, \texttt{Fuzzy}(c_i, s_j)\big]\Big).
\end{equation}
Within the self-attention layers, the prefix tokens dynamically query the rich, token-level semantics of the original fuzzy embeddings. We isolate the updated prefix tokens by extracting the first $N$ positions, denoted as $\mathbf{P}_{updated}^{(c_i, s_j)} = (\mathbf{C}'_{(c_i, s_j)})_{1:N}$. Finally, to preserve the initial generative intent while preventing gradient vanishing, we apply a residual connection followed by a terminal LayerNorm, yielding the final token templates
\begin{equation}
\mathcal{T}_{(c_i, s_j)} = \text{LayerNorm}\big(\mathbf{P}_{dyn}^{(c_i, s_j)} + \mathbf{P}_{updated}^{(c_i, s_j)}\big).
\end{equation}

If the diffusion model employs CLIP as a branch within a dual-stream or multi-stream architecture, parallel to the T5 branch,  we introduce a 2-layer Multilayer Perceptron (MLP) with a GELU activation function to process this pooled feature.
To ensure the projected global semantic anchor accurately captures the core semantics, we apply a Mean Squared Error (MSE) objective, to explicitly align these projected features with the target CLIP representation of the augmented prompt.
This term is ultimately integrated into the overall semantic alignment objective $\mathcal{L}_{\text{align}}$, as formulated in the main text.

\section{Details of Evaluation Metrics}
\label{sec:eval details}
\subsection{Details of Automated Metrics for Evaluating Creativity}
To comprehensively evaluate the  generated images across multiple dimensions—from semantic alignment to human aesthetic preference—we employ three widely recognized automated metrics: VQAScore, PickScore, and ImageReward.

\paragraph{VQAScore.} VQAScore~\cite{lin2024evaluating} is a metric designed to evaluate complex text-to-image alignment by formulating the evaluation as a Visual Question Answering (VQA) task. 
Given an image and a text prompt, it calculates the probability of a sophisticated Vision-Language Model answering ``Yes'' to the question ``Does this image accurately reflect the text prompt?''. 
Compared to traditional CLIP-based metrics, VQAScore demonstrates markedly stronger capabilities in capturing compositional prompts, precise attribute binding, and spatial relationships. 
These properties make it particularly suitable for evaluating creative generation, as it can more accurately reflect the nuanced interplay of multiple concepts, stylistic elements, and their spatial arrangements, which are key aspects of novelty and conceptual integration in creative outputs.

\paragraph{PickScore.} PickScore~\cite{kirstain2023pick} is a specialized metric trained to predict human choices and aesthetic preferences in text-to-image generation. It is built upon a CLIP-like architecture and fine-tuned on the large-scale Pick-a-Pic dataset, which contains hundreds of thousands of human pairwise preference annotations. Rather than solely measuring rigid semantic similarity, PickScore implicitly captures human aesthetic tastes, rendering quality, and naturalness, providing a holistic score that closely aligns with human preferences. 
Consequently, PickScore is well-suited for assessing creative generation, as it can capture subjective aspects of novelty, originality, and overall visual appeal in generated images.

\paragraph{ImageReward.} ImageReward~\cite{xu2024imagereward} serves as a comprehensive, general-purpose human preference reward model for text-to-image synthesis. Trained on a rigorously annotated dataset containing expert human rankings, ImageReward explicitly optimizes for multiple crucial dimensions of image generation. It actively penalizes common generative artifacts (such as distorted limbs or anatomical anomalies) while rewarding high-fidelity text alignment and visual appeal. It outputs a scalar reward value, where a higher score indicates a superior generation that aligns seamlessly with human structural and aesthetic expectations.
In this context, ImageReward effectively evaluates creative generation by capturing both fidelity to the prompt and overall visual coherence, reflecting qualities essential for assessing novelty, originality, and aesthetic appeal in generated outputs.

\subsection{Details of VLM-as-a-judge Evaluation}
\label{subsec:gpt prompt}
While automated metrics such as VQAScore, PickScore, and ImageReward offer valuable quantitative insights into semantic alignment and human preference, they remain limited in evaluating context-specific creativity and holistic conceptual integration in generated images.
To further quantitatively evaluate the generation quality and creativity of the synthesized images, we employ GPT-4o as a VLM judge. 
Recognizing that different categories require distinct evaluation standards, we carefully design task-specific prompts tailored for \textit{Architectural Design}, \textit{Furniture Design}, and \textit{Nature Mixture}. 
The VLM is prompted to function as an expert reviewer, evaluating each image across four key dimensions—Conceptual Integration, Prompt Alignment, Originality, and Aesthetic Quality—using a standardized 1–10 scoring rubric.

The complete prompts are as follows.
\begin{tcolorbox}[
    colback=blue!5!white,      
    colframe=blue!60!black,    
    title=\textbf{Prompt for GPT-4o Evaluation}, 
    arc=2mm,                   
    boxrule=0.5pt,             
    left=6pt, right=6pt, top=6pt, bottom=6pt,
    breakable 
]

% ================= 分支 1 =================

% ================= 分支 2 =================
\textbf{\large [Architecture Design]} \vspace{0.3em}

\texttt{[Role]} You are an extremely critical reviewer for a top-tier architectural design and generative AI conference.

\texttt{[System Instruction]} The subject of this evaluation is an image generated from the prompt: ``A creative [Target 2]-inspired [Target 1].'' The objective is to strictly assess the creativity and generation quality of an entity that achieves an organic fusion between the core of a [Target 1] and the inspiration of a [Target 2]. Evaluate the image utilizing the following criteria:

[EVALUATION CRITERIA]
\begin{enumerate}
    \setlength{\itemsep}{0pt}
    \setlength{\parskip}{0pt}
    \item (Conceptual Integration): Gauges the degree to which the inspiration is integrated into the architectural geometry. Must transcend mere surface texture wrapping.
    \item (Alignment with Prompt): Evaluates if the structure respects the macroscopic scale and architectural typology of a [Target 1], while vividly embodying the thematic traits of a [Target 2].
    \item (Originality): Innovativeness of the architectural silhouette and concept.
    \item (Aesthetic Quality): Visual realism, environmental lighting, and the absence of chaotic structural collapses or impossible physics.
\end{enumerate}

[CRITICAL SCORING RUBRIC (1-10)]
\begin{itemize}
    \setlength{\itemsep}{0pt}
    \setlength{\parskip}{0pt}
    \item [(1-3)]: Severe Failure. Structural collapse, unrecognizable as a [Target 1], or chaotic geometric mess.
    \item [(4-5)]: Subpar. Simplistic juxtaposition or mere 2D texture painting without structural modification.
    \item [(8-10)]: Excellent to Masterpiece. Highly organic structural fusion, breathtaking silhouette, pristine visual quality.
\end{itemize}

\vspace{1em} \hrule \vspace{1em} % 分割线

% ================= 分支 3 =================
\textbf{\large [Furniture Design]} \vspace{0.3em}

\texttt{[Role]} You are an extremely critical reviewer for a top-tier furniture design and generative AI conference.

\texttt{[System Instruction]} The subject of this evaluation is an image generated from the prompt: ``A creative [Target 2]-inspired [Target 1].'' The objective is to strictly assess the creativity and generation quality of an entity that achieves an organic fusion between the core of a [Target 1] and the inspiration of a [Target 2]. Evaluate the image utilizing the following criteria:

[EVALUATION CRITERIA]
\begin{enumerate}
    \setlength{\itemsep}{0pt}
    \setlength{\parskip}{0pt}
    \item (Conceptual Integration): Gauges how well the inspiration is woven into the physical structure of the furniture. Must organically alter the structural language without looking like a cheap surface sticker or illogical attachment.
    \item (Alignment with Prompt): MUST strictly retain the core utilitarian purpose of a [Target 1] (avoid 'Functional Deprivation'). It must seamlessly adopt the aesthetic traits of a [Target 2] while remaining a physically viable piece of furniture.
    \item (Originality): Innovativeness of the furniture silhouette and stylistic amalgamation. Does it break conventional molds while remaining recognizable?
    \item (Aesthetic Quality): Rendering of furniture-specific materials, natural lighting interactions, and the absence of AI structural artifacts (e.g., melting legs, asymmetrical bases causing imbalance).
\end{enumerate}

[CRITICAL SCORING RUBRIC (1-10)]
\begin{itemize}
    \setlength{\itemsep}{0pt}
    \setlength{\parskip}{0pt}
    \item [(1-3)]: Severe Failure. Functional Deprivation (loses basic utility) or crude cut-and-paste juxtaposition.
    \item [(4-5)]: Subpar. Retains basic furniture function, but fusion is forced, ergonomically illogical, or suffers from severe artifacts.
    \item [(8-10)]: Excellent to Masterpiece. Brilliant furniture design, ergonomically sound, highly functional yet creatively morphed, and visually pristine.
\end{itemize}

\vspace{1em} \hrule \vspace{1em} % 分割线

\textbf{\large [Nature Mixture]} \vspace{0.3em}

\texttt{[Role]} You are an extremely critical reviewer for a top-tier generative AI and digital biology conference.

\texttt{[System Instruction]} The subject of this evaluation is a creative mixture image in nature domain. The objective is to strictly assess the creativity and generation quality of an entity that achieves an organic fusion between a [Target 1] and a [Target 2]. Evaluate the image utilizing the following criteria:

[EVALUATION CRITERIA]
\begin{enumerate}
    \setlength{\itemsep}{0pt}
    \setlength{\parskip}{0pt}
    \item (Conceptual Integration): Are the elements organically grown together? A high score signifies anatomical harmony.
    \item (Alignment with Prompt): Does it retain the recognizable life-form identity of a [Target 1] while vividly exhibiting the genetic/stylistic traits of a [Target 2]?
    \item (Originality): Innovativeness of the creature design.
    \item (Aesthetic Quality): Visual appeal, natural textures (fur, scales, skin), and lack of disturbing anatomical artifacts.
\end{enumerate}

[CRITICAL SCORING RUBRIC (1-10)]
\begin{itemize}
    \setlength{\itemsep}{0pt}
    \setlength{\parskip}{0pt}
    \item [(1-3)]: Severe Failure. Concept deprivation, disturbing anatomical anomalies (extra limbs), or crude cut-and-paste.
    \item [(4-5)]: Subpar. Traits are somewhat blended but biologically nonsensical or suffer from visual artifacts.
    \item [(8-10)]: Excellent to Masterpiece. Highly organic biological fusion, visually pristine.
\end{itemize}

\vspace{1em} \hrule \vspace{1em} % 分割线

% ================= 公共输出要求 =================
\textbf{\large [Common Instructions for All Domains]} \vspace{0.3em}

[ANTI-BIAS INSTRUCTION] \\
Do NOT default to safe, average scores (6-7). Aggressively penalize (1-5 points) crude pasting, loss of structural integrity/function, or severe artifacts. Reward exceptional organic fusion with scores of 8 or above.

[OUTPUT INSTRUCTION] \\
You MUST output strictly as a JSON object. CRITICAL: To ensure a rigorous evaluation, you MUST output the ``Justification'' key FIRST. In the justification, explicitly search for flaws, missing concepts, or crude blending. ONLY THEN output the 5 numeric scores.

Exact keys required in this order: ``Justification'', ``Conceptual\_Integration'', ``Alignment'', ``Originality'', ``Aesthetic\_Quality'', ``Comprehensive\_Assessment''. The values for the last 5 keys MUST be integers or floats between 1 and 10.
\end{tcolorbox}

In Table~\ref{tab:gpt4o_eval}, we present detailed sub-item scores from the VLM-as-a-judge evaluation. 
Our method attains state-of-the-art performance across nearly all evaluated dimensions, demonstrating its superior capability in integrating concepts, aligning with prompts, and producing visually appealing and original creativity.

\begin{table}[h]
    \centering
    \setlength{\tabcolsep}{2mm} 
    \caption{\textbf{Detailed Results of VLM-as-a-judge.} We reported detailed sub-scores for Integration, Alignment, Originality, and Aesthetic.}
    \label{tab:gpt4o_eval}
    \small 
    \resizebox{0.9 \textwidth}{!}{
    \begin{tabular}{@{}cl cccc|c@{}}
    \toprule[1.3pt]
    & \textbf{Method} & \textbf{Integration} & \textbf{Alignment} & \textbf{Originality} & \textbf{Aesthetic} & \textbf{\textit{Comprehensive}} \\
    \midrule[1.3pt]
    \multirow{3}{*}{\rotatebox[origin=c]{90}{\textbf{Arch.}}} 
    % & Flux.1   & 0.00 & 0.00 & 0.00 & 0.00 & 0.00 \\
    & T2I-Copilot & 6.80 $\pm$ 1.95 & 7.25 $\pm$ 1.94 & 6.97 $\pm$ 1.69 & 8.43 $\pm$ 0.67 & \textit{7.30 $\pm$ 1.60} \\
    & CREA        & 7.60 $\pm$ 1.89 & 7.60 $\pm$ 1.70 & \textbf{8.03 $\pm$ 1.42} & 8.65 $\pm$ 0.75 & \textit{7.82 $\pm$ 1.39} \\
    & \cellcolor{blue!12}{\textbf{CAT}} & \cellcolor{blue!12}{\textbf{8.45 $\pm$ 1.07}} & \cellcolor{blue!12}{\textbf{8.60 $\pm$ 0.82}} & \cellcolor{blue!12}{7.83 $\pm$ 0.99} & \cellcolor{blue!12}{\textbf{8.85 $\pm$ 0.49}} & \cellcolor{blue!12}{\textbf{\textit{8.44 $\pm$ 0.80}}} \\
    \midrule
    
    \multirow{3}{*}{\rotatebox[origin=c]{90}{\textbf{Furni.}}} 
    % & Flux.1  & 0.00 & 0.00 & 0.00 & 0.00 & 0.00 \\
    & T2I-Copilot & 7.92 $\pm$ 1.64 & 8.10 $\pm$ 1.52 & 7.55 $\pm$ 1.36 & 8.50 $\pm$ 1.05 & \textit{7.88 $\pm$ 1.23} \\
    & CREA        & 7.78 $\pm$ 1.44 & 7.20 $\pm$ 1.70 & 7.95 $\pm$ 1.65 & 8.15 $\pm$ 0.67 & \textit{7.64 $\pm$ 1.17} \\
    & \cellcolor{blue!12}{\textbf{CAT}} & \cellcolor{blue!12}{\textbf{8.22 $\pm$ 1.71}} & \cellcolor{blue!12}{\textbf{8.18 $\pm$ 1.16}} & \cellcolor{blue!12}{\textbf{8.07 $\pm$ 1.55}} & \cellcolor{blue!12}{\textbf{8.50 $\pm$ 0.89}} & \cellcolor{blue!12}{\textbf{\textit{8.20 $\pm$ 1.21}}} \\
    \midrule
    
    \multirow{4}{*}{\rotatebox[origin=c]{90}{\textbf{Nature}}} 
    % & Flux.1  & 0.00 & 0.00 & 0.00 & 0.00 & 0.00 \\
    & BASS        & 6.50 $\pm$ 2.56 & 6.73 $\pm$ 2.13 & 6.97 $\pm$ 2.33 & 7.43 $\pm$ 1.82 & \textit{6.81 $\pm$ 2.24} \\
    & AGSwap      & 7.33 $\pm$ 2.25 & 7.17 $\pm$ 2.21 & 7.53 $\pm$ 2.11 & 8.02 $\pm$ 1.51 & \textit{7.50 $\pm$ 2.03} \\
    & CreTok      & 7.40 $\pm$ 2.24 & 7.35 $\pm$ 2.00 & 7.48 $\pm$ 1.74 & 7.92 $\pm$ 1.66 & \textit{7.59 $\pm$ 1.93} \\
    & \cellcolor{blue!12}{\textbf{CAT}} & \cellcolor{blue!12}{\textbf{8.83 $\pm$ 0.77}} & \cellcolor{blue!12}{\textbf{8.78 $\pm$ 0.69}} & \cellcolor{blue!12}{\textbf{8.38 $\pm$ 0.58}} & \cellcolor{blue!12}{\textbf{8.85 $\pm$ 0.48}} & \cellcolor{blue!12}{\textbf{\textit{8.82 $\pm$ 0.56}}} \\
    \bottomrule[1.3pt]
    \end{tabular}
    }
\end{table}

\section{Details of the Dataset}
\label{sec:data details}
We extend the original \textit{CangJie} dataset to encompass creative design tasks, thereby enhancing the practicality of creative generation. 
Table~\ref{tab:dataset_overview} presents the complete contents of the expanded \textit{CangJie} dataset~\cite{feng2025redefining}, including all primary and style-guiding concepts across diverse design and concept mixture settings, providing a comprehensive basis for evaluating compositional capabilities.

\begin{table*}[h]
    \centering
    \caption{\textbf{Overview of the Expanded \textit{CangJie} Dataset.} 
    For the \textit{Architecture Design} and \textit{Furniture Design} tasks, each pair consists of a primary concept and a style-guiding concept.  
    For the \textit{Nature Mixture} task, the two concepts in each pair are combined in equal proportion.
    All items are listed in alphabetical order.}

    \label{tab:dataset_overview}
    \begin{tabularx}{\textwidth}{@{} l >{\raggedright\arraybackslash\hspace{0pt}}X >{\raggedright\arraybackslash\hspace{0pt}}X @{}}
        \toprule[1.3pt] 
        \textbf{Task} & \textbf{Concept} & \textbf{Style} \\
        \midrule 

        % ====== Architecture ======
        \textbf{Architecture Design} & 
        \textbf{(10 items):} Clock Tower, Museum, Observation Tower, Observatory, Pavilion, Shopping Mall, Skyscraper, Stadium, Suspension Bridge, Windmill & 
        \textbf{(55 items):} Accordion, Armadillo Shell, Bamboo, Bat Wing, Bird Nest, Bookshelf, Bubble, Butterfly, Cactus, Camera Lens, Candle, Caterpillar, Circuit Board, Coral Reef, Corn, Crystal, Dandelion, Deer Antler, DNA Double Helix, Dolphin, Fish, Folding Fan, Gyroscope, Honeycomb, Hourglass, Jellyfish, Lantern, Leaves, Lotus, Mountain, Mushroom, Onion, Origami Crane, Peacock Feather, Peanut, Piano, Pinecone, Pipe Organ, Pocket Watch, Prism, Ribcage, Rose, Sailing Ship, Sea Urchin, Seashell, Shark, Snail Shell, Spider Web, Stained Glass, Sunflower, Tortoise Shell, Tree Branches, Vinyl Record, Walnut, Woven Basket \\
        
        \addlinespace
        \midrule
        \addlinespace
        
        % ====== Furniture ======
        \textbf{Furniture Design} & 
        \textbf{(20 items):} Bar Stool, Bed, Bookshelf, Cabinet, Chair, Chandelier, Coat Rack, Coffee Table, Cradle, Dining Table, Dresser, Fireplace, Floor Lamp, Mirror, Sofa, Swing, Table Lamp, Television, Tub, Vase & 
        \textbf{(38 items):} Accordion, Armadillo Shell, Bamboo, Bat Wing, Bird Nest, Bubble, Butterfly, Cactus, Circuit Board, Coral Reef, Crystal, Dandelion, Deer Antler, Folding Fan, Honeycomb, Hourglass, Jellyfish, Lantern, Leaves, Lotus, Mushroom, Origami Crane, Peacock Feather, Pinecone, Prism, Ribcage, Rose, Sea Urchin, Seashell, Snail Shell, Spider Web, Stained Glass, Sunflower, Tortoise Shell, Tree Branches, Vinyl Record, Walnut, Woven Basket \\
        
        \addlinespace
        \midrule[1.3pt]
        \addlinespace
        
        % ====== Nature Mixture ======
        \textbf{Task} & \multicolumn{2}{l}{\textbf{Concept}} \\
        \midrule
        \textbf{Nature Mixture} & 
        \multicolumn{2}{>{\raggedright\arraybackslash}p{\dimexpr0.78\textwidth\relax}}{%
            \textbf{(60 items):} Ant, Antelope, Bat, Bear, Bee, Broccoli, Butterfly, Cabbage, Cactus, Camel, Carrot, Cat, Cauliflower, Chameleon, Cherry, Chicken, Coconut, Crab, Crocodile, Deer, Dog, Dolphin, Eagle, Elephant, Fox, Frog, Giraffe, Gorilla, Grape, Hippo, Horse, Jellyfish, Kangaroo, Lemon, Lettuce, Lion, Mango, Mushroom, Octopus, Owl, Peacock, Penguin, Persimmon, Pineapple, Potato, Pumpkin, Rabbit, Rhino, Snail, Snake, Spider, Starfruit, Strawberry, Sunflower, Tiger, Tomato, Turtle, Watermelon, Whale, Zebra
        } \\
        \bottomrule[1.3pt] 
    \end{tabularx}
\end{table*}

\section{Details of the User Study}
\label{sec:app_user}
\begin{figure}[h]
  \centering
  \includegraphics[width=0.9 \linewidth]{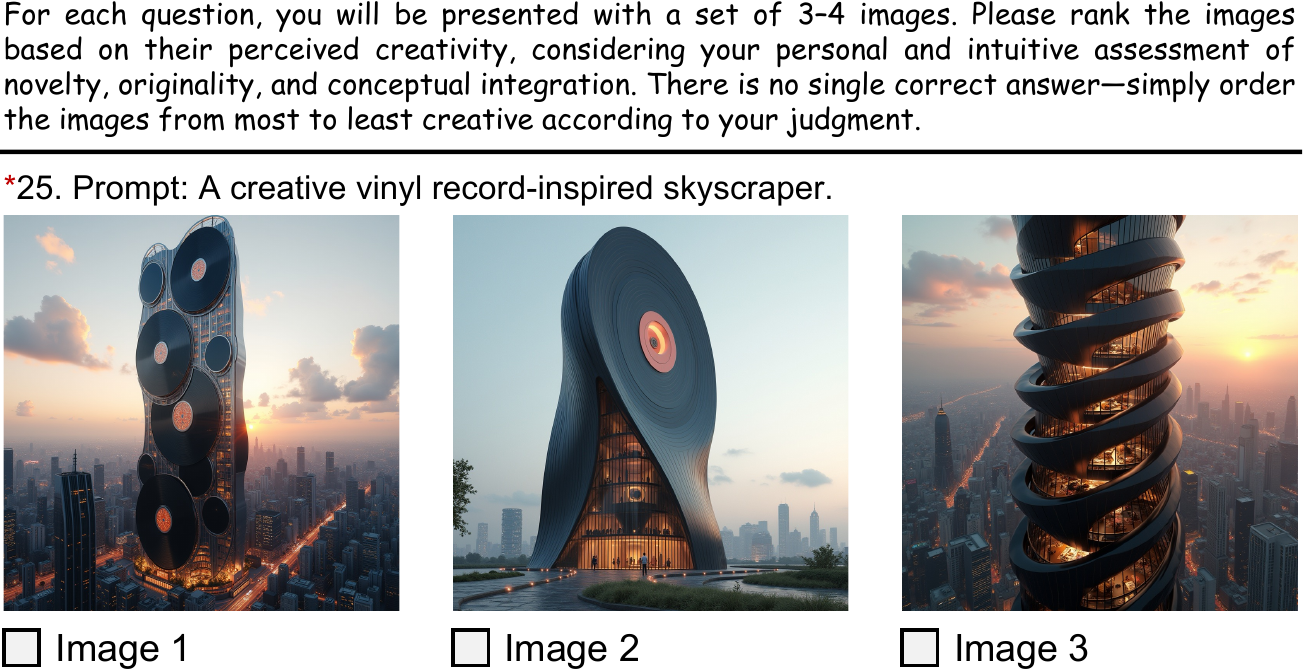}
  \vspace{-0.05in}
  \caption{\textbf{Interface of the User Study.} 
  Each volunteer is assigned a subset of questions from three tasks: \textit{Architecture Design} (20 questions), \textit{Furniture Design} (20 questions), and \textit{Nature Mixture} (30 questions). 
  For each participant, 10 questions are randomly selected from each task. 
  For each question, participants rank the presented images based on perceived creativity.}
\label{fig:app_infer}
\end{figure}

To obtain a comprehensive and balanced evaluation of visual creativity, we seek 30 volunteers with diverse expertise, encompassing both professional and non-professional backgrounds in art, design, and visual cognition. 
This participant diversity allows us to assess the generated concepts from multiple perspectives, capturing both expert-level criteria and general audience preferences. 
The study interface is presented in Fig.~\ref{fig:app_infer}.

% \section{Supplementary Experiments}
%   \subsection{Analysis of VLM used in Creative Agent System}
%    \subsection{Rejection Mechanism of Creative Evaluator}

\section{Additional Results}
\begin{figure}[tb]
  \centering
  \includegraphics[width=0.9\linewidth]{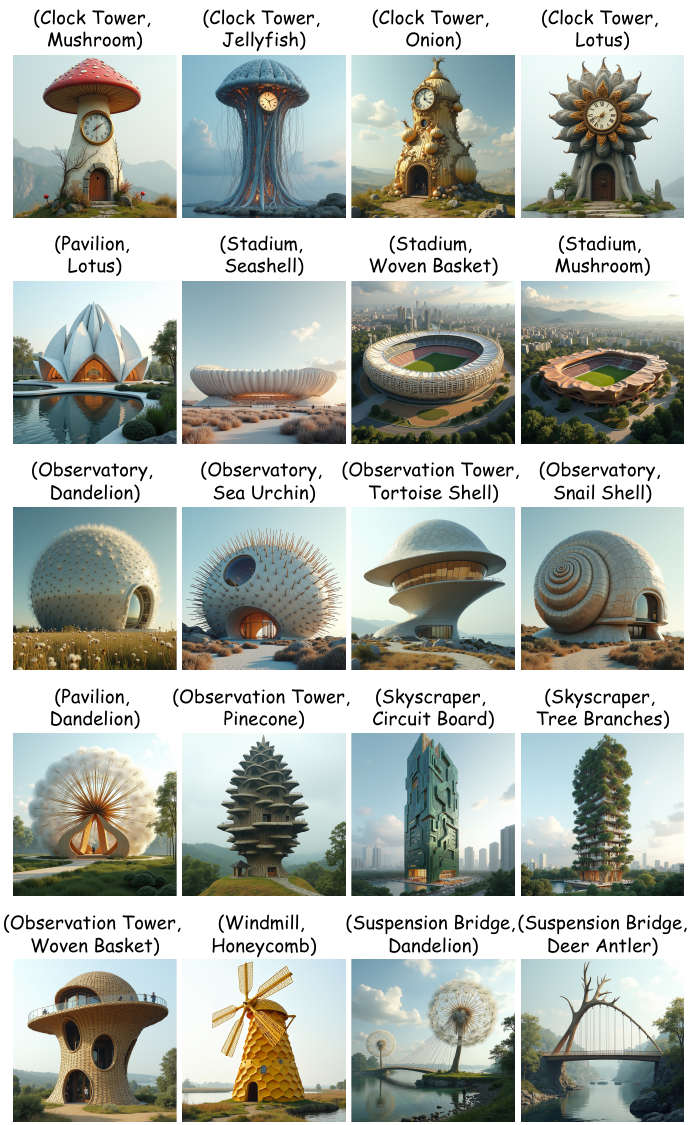}
  \caption{\textbf{Qualitative results of CAT on the \textit{Architecture Design} task.}} 
  \label{fig:app_arch}
\end{figure}

\begin{figure}[tb]
  \centering
  \includegraphics[width=0.9\linewidth]{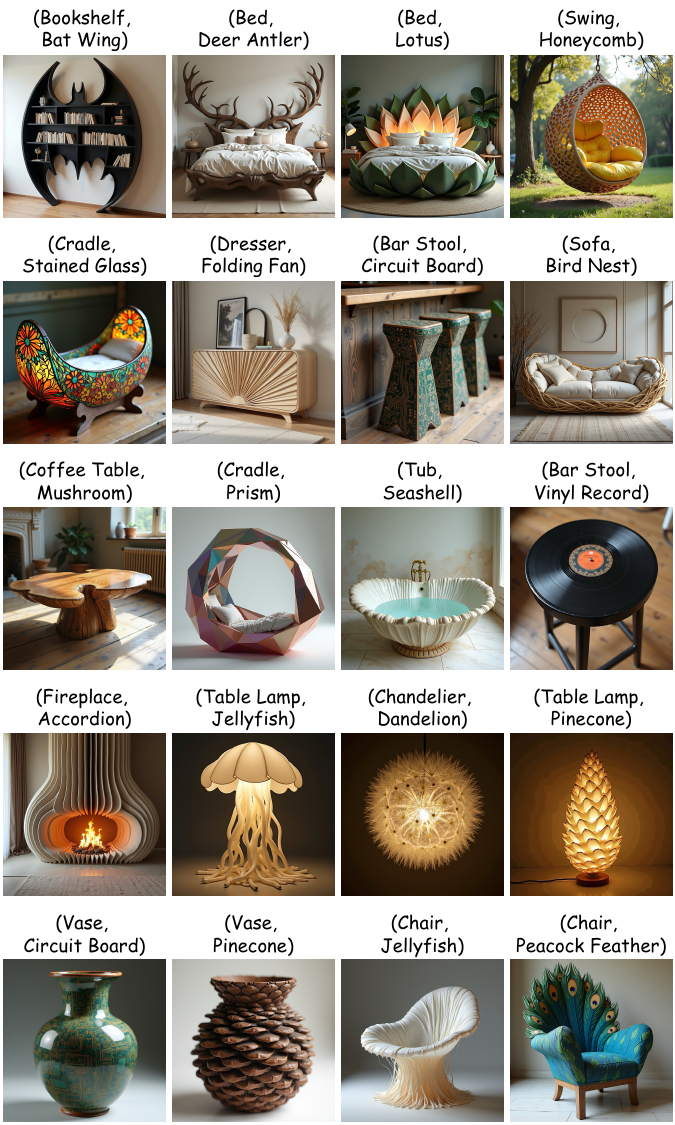}
  \caption{\textbf{Qualitative results of CAT on the \textit{Furniture Design} task.}} 
  \label{fig:app_furn}
\end{figure}

\begin{figure}[tb]
  \centering
  \includegraphics[width=0.9\linewidth]{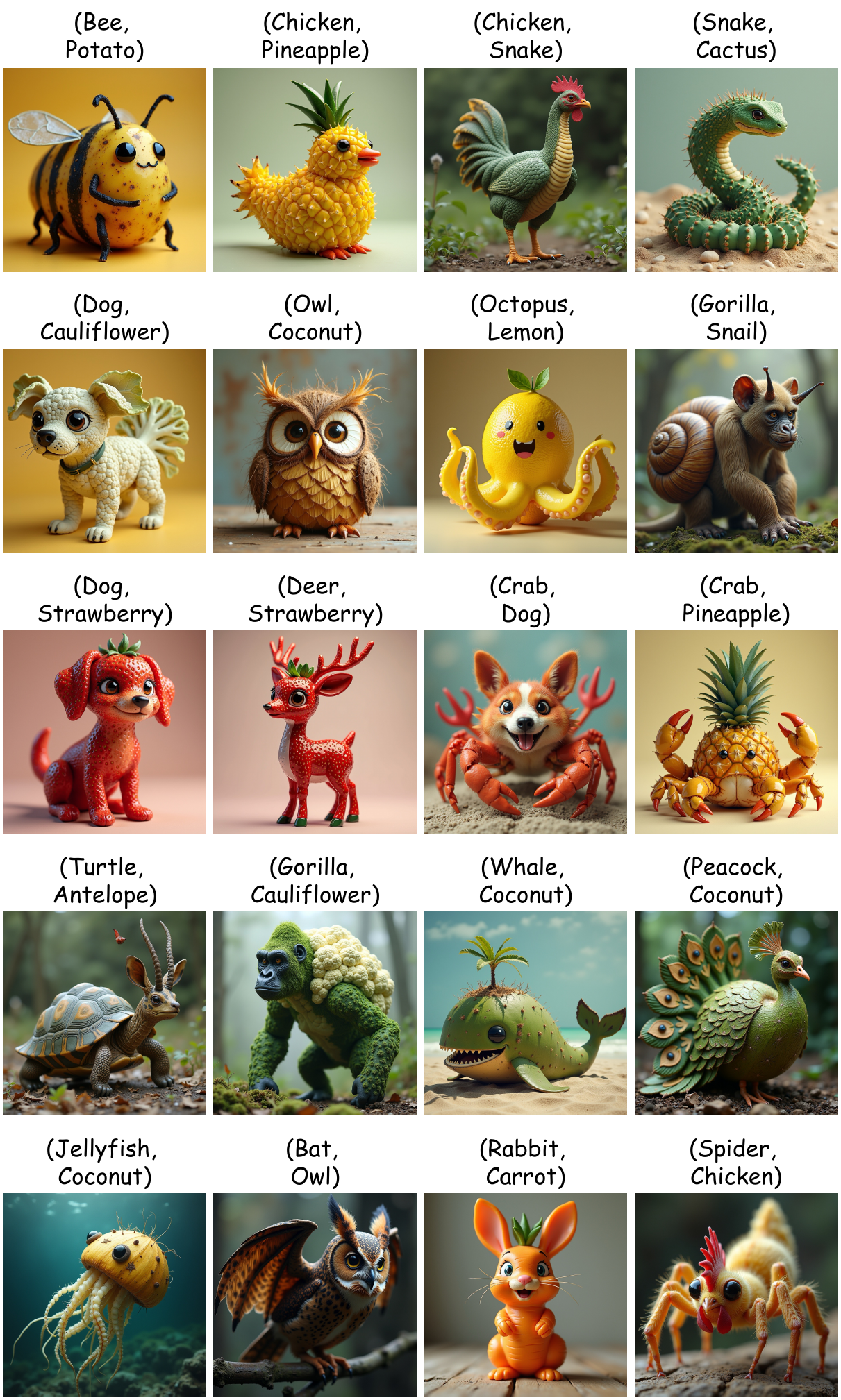}
  \caption{\textbf{Qualitative results of CAT on the \textit{Nature Mixture} task.}} 
  \label{fig:app_nature}
\end{figure}

To further illustrate the performance of CAT across different creative generation domains, we provide additional qualitative results for the \textit{Architecture Design}, \textit{Furniture Design} and \textit{Nature Mixture} task in Fig.~\ref{fig:app_arch}–\ref{fig:app_nature}. 
These examples highlight CAT's ability to generate coherent and aesthetically appealing compositions while maintaining conceptual and stylistic integrity.

\section{Limitations and Future Work}
While CAT exhibits strong generative capabilities, it currently depends on VLMs, such as GPT-4o, for both prompt augmentation and evaluation. 
Since the Creative Tokenizer learns from these VLM-derived signals, the diversity and stylistic characteristics of the generated outputs are inherently shaped by the model's knowledge base and aesthetic biases. 
As a result, any intrinsic preferences of the VLM—such as favoring conventional structural patterns—may propagate into the creative embeddings, potentially constraining fully open-ended creativity.

To further expand creative potential and reduce VLM-specific biases, future work will investigate direct preference alignment. 
By incorporating human feedback into the optimization process, this approach aims to better capture authentic human aesthetic judgment, structural fidelity, and functional ergonomics. 
We expect that such a preference-driven framework will advance generated concepts beyond mere semantic fusion, enabling more open-ended and human-aligned creative designs.

\end{document}